\documentclass[10pt,twocolumn,letterpaper]{article}

\usepackage{iccv}
\usepackage{times}
\usepackage{epsfig}
\usepackage{graphicx}
\usepackage{amsmath}
\usepackage{amssymb}
\usepackage[table,xcdraw]{xcolor}
\usepackage{caption}
\usepackage{authblk}
\usepackage{float}
\usepackage{xcolor}
\usepackage{multirow}

\usepackage[breaklinks=true,bookmarks=false]{hyperref}

\iccvfinalcopy

\ificcvfinal\fi

\begin{document}
\title{Road images augmentation with synthetic traffic signs using neural networks}

\makeatletter
\renewcommand*{\Authands}{, }
\makeatother
\author[1,2]{Anton Konushin}
\author[1, *]{Boris Faizov}
\author[1]{Vlad Shakhuro}
\makeatletter
\renewcommand\AB@affilsepx{,\ \ \ \ \ \ \ \ \  \protect\Affilfont}
\makeatother
\affil[1]{Lomonosov Moscow State University}
\makeatletter
\renewcommand\AB@affilsepx{,\\\protect\Affilfont}
\makeatother
\affil[2]{NRU Higher School of Economics}
\makeatletter
\renewcommand\AB@affilsepx{\\\protect\Affilfont}
\makeatother
\affil[ ]{\textit {\{firstname.lastname\}@graphics.cs.msu.ru}}
\affil[*]{Corresponding author}
\date{}

\maketitle

\begin{abstract}
Traffic sign recognition is a well-researched problem in computer vision.
However, the state of the art methods works only for frequent sign classes, which are well represented in training datasets. We consider the task of rare traffic sign detection and classification.  We aim to solve that problem by using synthetic training data. Such training data is obtained by embedding synthetic images of signs in the real photos. We propose three methods for making synthetic signs consistent with a scene in appearance. These methods are based on modern generative adversarial network (GAN) architectures. Our proposed methods allow realistic embedding of rare traffic sign classes that are absent in the training set. We adapt a variational autoencoder for sampling plausible locations of new traffic signs in images. We demonstrate that using a mixture of our synthetic data with real data improves the accuracy of both classifier and detector.
\end{abstract}

\section{Introduction}

\begin{figure}[h]
\begin{center}
\setlength\tabcolsep{3pt}
\renewcommand{\arraystretch}{1}
\begin{tabular}{cc}
\includegraphics[width=38mm]{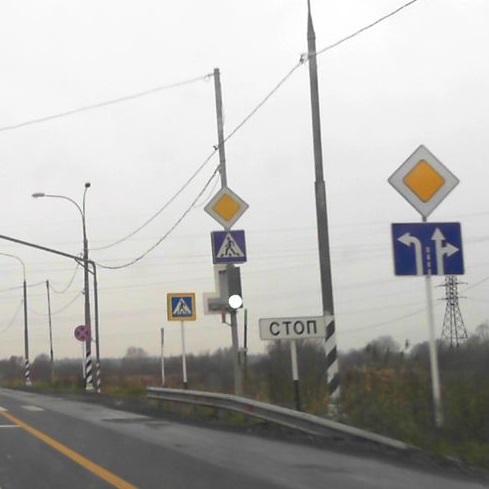} &
\includegraphics[width=38mm]{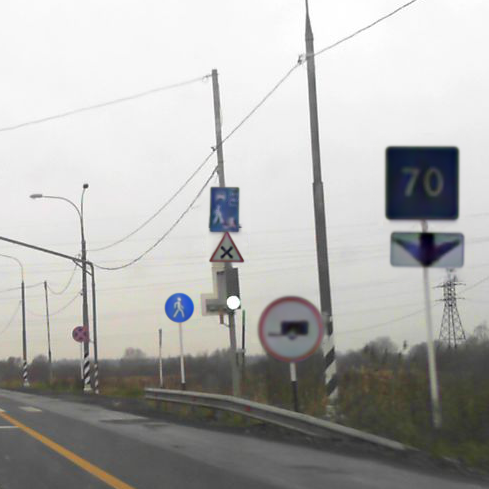} \\

(a) original image & (b) real signs replaced \\
\ & \ with synthetic \\
\ & \ \\

\includegraphics[width=38mm]{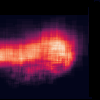} &
\includegraphics[width=38mm]{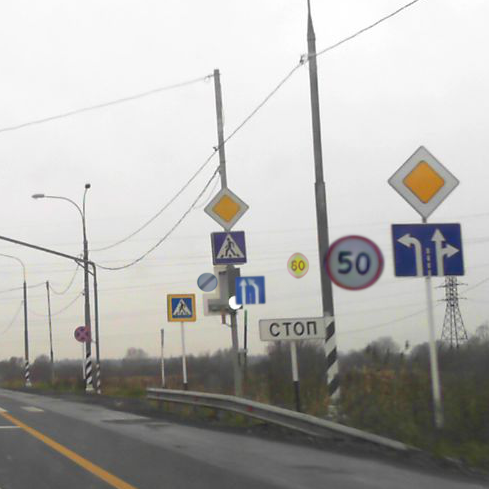} \\

(c) predicted sign & (d) additional synthetic \\
\ location heatmap & \ signs \\
\end{tabular}
\end{center}\caption{Example of fragment with 6 traffic signs. Here on one fragment real ones are replaced with new synthetic. On another fragment there are embedded new signs} \label{fig:example1}
\end{figure}

Modern computer vision methods are based on machine learning techniques and require labelled datasets for training. The accuracy of the trained model depends on the size and quality of the available dataset. Dataset labelling is a labor-consuming and time-consuming process that is prone to errors. In contrast, synthetic data generation can produce virtually unlimited training datasets without annotation errors. This is why methods for generating synthetic images are actively investigated in recent years. 

In this paper, we consider the task of generating artificial data for training traffic sign recognition models. Traffic sign recognition is a significant problem which gains the stable interest of researchers over the years. Traffic sign detection and classification are used in driver assistance systems, self-driving cars, for maintaining up-to-date high-resolution maps and traffic sign inventory. Modern open datasets for traffic sign recognition can contain thousands of frames with two hundred classes. However, a distinctive feature of the traffic sign recognition problem is a significant amount of rare classes. Objects of such classes can either be present in small amounts in datasets or absent. But it is still required to train recognition algorithms for such traffic sign classes since the importance of rare classes on the road is no less than that of frequent.

We investigate modern methods for generating synthetic training data using neural networks.  Since even state-of-the-art methods are unable to generate the whole photos of the traffic scene with photo-realistic quality, we propose to embed artificial signs in real images. Two questions arise immediately: how to make the inserted object consistent in appearance with the scene and where to position it.

We focus on the recognition of rare traffic sign classes. Since such signs are absent or limited in the real dataset, we can't directly train a neural network to generate images of such signs. Instead, we aim to create a synthetic traffic sign processing method that will improve the realism of simple synthetic images obtained from the sign icon. We propose three processing methods based on generative adversarial networks~\cite{ganbasic,cyclegan,stylegan}.  We embed artificial signs in reals images instead of already existing traffic signs. To do this, we first remove the existing ones via inpainting and then place synthetic signs on their places (see Figure \ref{fig:example1}b).  Inpainting is done using a neural network that is trained separately or jointly with the sign processing method. Such a technique allows us to augment images with rare sign classes with the correct geometric placement and evaluate the individual contribution of object processing methods.

In the second part of our work, we adopt a method based on variational autoencoder \cite{placementcaware} to predict the correct location and size for synthetic traffic sign insertion. To predict plausible traffic sign placement in a frame, we first automatically obtain semantic segmentation of the image and then sample locations using variation autoencoder. An example of the obtained heatmap is shown in Figure \ref{fig:example1}c. After obtaining locations, we insert synthetic traffic signs in addition to real traffic signs in a frame (see Figure \ref{fig:example1}d).

Overall, we propose three methods for processing synthetic traffic signs and a new method for their placement on real road images at the geometrically correct position. Proposed methods allow augmenting the real road images with high-quality synthetic traffic signs for classes, which are absent in the real training dataset. We have conducted an extensive experimental evaluation of the proposed methods. It has demonstrated that usage of generated data improves the quality of traffic sign detection and classification, especially for the rare classes.

\section{Related work}
\subsection{Synthetic image generation and processing}
The augmentation of real images with new synthetic objects can be implemented using different methods. The simplest and most obvious way is to draw an object without any processing \cite{dwibedi2017cut, dvornik2018modeling}. However, this approach will lead to unrealistic images and does not allow to obtain high-quality synthetic samples. Recently generative adversarial neural networks \cite{ganbasic} have been applied to such problems.  Such methods perform image processing so that artificial objects matches the background in colour and lighting \cite{zhang2019shadowgan, liu2019generative, reed2019coupling}. However, the geometric position and shape of the embedded objects are still not taken into account. 

The basic idea of a generative adversarial network is to have two separate parts -- a generator and a discriminator. The generator creates synthetic images. Discriminator learns to distinguish generated images from real ones. These neural network’s components try to deceive each other during the training process. In \cite{radford2015unsupervised}, for image generation, convolutional neural network architecture with transposed convolutions was proposed to increase the resolution of generated images. The proposed approach with convolutional layers made it possible to train the neural network faster and improve the quality. Other authors \cite{denton2015deep} used the Laplace pyramid and several generators and discriminators. Also, the researchers proposed work on the conditional generation of an object with a given class \cite{mirza2014conditional}. The generator receives not only random noise at the input but also the class label of the object to generate.

GAN models have been successfully applied for image transfer between domains. One of the notable examples is CycleGAN \cite{cyclegan}, which doesn't need labelled pairs of images from the source and target domains for training.  It has two generators and two discriminators. Suppose we have two image domains $\mathcal{A}$ and $\mathcal{B}$. The first generator learns to transfer images from $\mathcal{A}$ to $\mathcal{B}$, and the second generator on the contrary from $\mathcal{B}$ to 
$\mathcal{A}$. First discriminator trains to distinguish synthetic and real
images from $\mathcal{B}$, and the second discriminator vice versa. During the inference process, only the desired generator is used. 

The rapid progress of GANs is quite astonishing. In 2019 a StyleGAN architecture was proposed \cite{stylegan}, which demonstrates a surprisingly realistic generation of people's faces. They were generated from random vectors, which were at first transformed by a small fully-connected part of a neural network to obtain a vector in intermediate latent space. The Adaptive Instance normalization (AdaIN) layers \cite{adain} are used in the generator to transfer information from vector in latent space. Also, random noise is actively added to the architecture in the intermediate layers to obtain a variety in the small details of the individual generated images. Our proposed methods for high-quality synthetic traffic sign generation are based on this approach. 

The previous methods don't predict the location of embedded objects. In the paper \cite{cocacolaplacement}, the authors suggested the adversarial approach for generating synthetic object placement and processing. A proposed neural network has a branch for predicting the location and size of a new object. A simple colour correction of six predicted parameters is used for the first stage of object processing. Then a refinement network is used to improve object
consistency with the background. As usual, architecture has discriminator for distinguishing synthetic image and new segmentation network which learns to predict a mask of the artificial object.

The usage of synthetic training data for accuracy improvement of recognition models is actively investigated. In \cite{zheng2017unlabeled}, the quality of the re-identification of people in video was improved by adding synthetic data to real data. In \cite{frid2018synthetic}, synthetic data were added to the training set to improve the quality of liver lesions classification. In papers \cite{richter2016playing, gaidon2016virtual}, game engines are used to generate the labelled city scenes. Synthetic data made it possible to improve the quality of the final algorithm and reduce the  requirements for the amount of real data by three times.

The authors of paper \cite{hoffman2018cycada} suggested generating synthetic road images from the GTA computer game by transferring data from one domain to another. As a target domain, they used images from the Cityscapes \cite{cordts2016cityscapes} dataset. Their approach is
based on CycleGAN architecture.

\subsection{Predicting synthetic object locations}

Most modern neural network architectures for changing position and parameters of objects are based on Spatial Transformer Networks \cite{stn}. The idea of such architectures is to add a separate part of a neural network that will generate affine transformation parameters for an added object according to the background. The resulting affine transformation is applied to a grid of pixels that specify where and which pixel of an object will be positioned. The article shows that these transformations are differentiable and can be optimized in neural
networks.

Spatial Transformer Networks became popular for synthetic object placement. Authors of works \cite{cocacolaplacement, compositingstn} based their approaches for generating locations of new entities by predicting affine transformations. Besides discriminator during the training process, these methods relied on an additional signal such as image segmentation network or target classifier/detector networks. Disadvantages of a spatial transformer are bad convergence, instability, and complex training process.

Article \cite{placementcaware} proposed a VAE-based approach for object placement on road images. The algorithm has two separate modules for determining \emph{where} and \emph{what} could be placed in the picture. A generator with Spatial Transformer Networks is used in the first module to determine where to place the object. In the second module, there is a generator for the shape of an embedded object to determine what exactly needs to be placed. 

\subsection{Traffic sign generation and recognition}
The traffic sign recognition methods have a long history.  Early approaches were based on finding corners and feature points in images \cite{de1997road}. Usage of synthetic training datasets has been investigated since 2007 \cite{hoessler2007classifier}. Generation of synthetic examples for training traffic sign classifiers has been implemented by applying affine transformations to sign pictograms.

In \cite{simpleisignrecognition} a four-stage system for detection and classification of traffic signs was proposed. It included a cascade detector and a set of neural network classifiers for each type of traffic sign. This model was trained with synthetic data proposed in paper \cite{earlyrtsdsynt}. A suggested approach for the generation of synthetic data used heuristic methods, based on computer graphics. But it also tried to predict the best parameters for current data. Paper showed that models trained on synthetically generated data could produce good results.

Since the introduction of modern deep learning methods, they have been applied to the traffic sign recognition. In the paper \cite{zhu2016traffic}, the authors first collected a massive training dataset and then proposed a fully convolutional simple neural network architecture for the simultaneous detection and classification of traffic signs.

In \cite{icviprtsd} synthetic traffic signs with poles were generated by computer graphics methods and then improved with a neural network, base on CycleGAN. To preserve the traffic sign class while processing an additional identity loss has been using during the training of this model. Then artificial traffic signs were embedded into reals photos. Simple heuristics based on a reconstruction of camera parameters and simple 3D-modelling was used to determine new signs location. Experimental evaluation showed that this approach works better compared to random object placement. However, the best results were obtained if new artificial traffic signs replace existing ones. 

Currently, the best results in traffic sign recognition are achieved by adapting modern detection architectures. For example, in \cite{liang2019traffic, ayachi2020traffic} anchor-based methods have been used, with specific optimizations for speed. In \cite{liang2019traffic} authors used the ResNet-50 as the backbone to build a pyramidal feature network. In \cite{ayachi2020traffic} MobileNet-backbone with suggested Localization network is used.

Conventional convolutional neural networks can be used for traffic signs classification, such as AlexNet\cite{krizhevsky2017imagenet}, ResNet\cite{he2016deep} etc. In 2019, the article \cite{farag2019traffic} proposed to apply the special pre-processing procedure to the traffic signs before classification. Then, processed signs are fed to the input of a small neural network, based on LeNet.

Different convolutional architectures were tested on the traffic sign classification problem in the article \cite{serna2018classification}. Authors showed that CNN architectures are well suited for this task and drew attention to the lack of real data. The authors concluded that techniques like image pre-processing and data-augmentation are useful to improve classification accuracy.

Another approach aimed at solving the problem of rare traffic signs classification was proposed in 2020 \cite{betterwideresnet}. Authors use WideResNet \cite{wideresnet}, trained with contrastive loss, for feature extraction. One discriminator is used to distinguish rare and frequent classes. Authors proposed to classify frequent classes using features from the last layer of the neural network, and rare classes using the nearest neighbour method.

\section{Proposed methods}
We explored two different ways to embed traffic signs in images:
\begin{itemize}
    \item \textbf{Replacement of existing real traffic sings with artificial ones}. In this case, we use inpainting at the place of the real sign to generate plausible background. Then artificial traffic sign is embedded on top of it.  This way of generating synthetic data allows increase training set with new examples of rare classes with the correct geometric position. The article  \cite{icviprtsd} showed that this approach improves the quality of neural networks for classification and detection. It also allows us to evaluate better the individual contribution of proposed processing methods for improving target neural networks quality. For inpainting we used Edgeconnect \cite{edgeconnect} architecture.
    \item \textbf{ Embedding additional artificial signs in new positions.} In this case, we need to learn how to find the most suitable position for the new traffic signs first, and then perform their processing. To find the correct position of new traffic signs, a neural network architecture based on \cite{placementcaware} was chosen.
\end{itemize}

\subsection{Processing of embedded traffic signs}
In both ways, we need to process artificial signs to improve visual consistency with the background. We propose three models for this task. The first two of them are trained together with the inpainting network, and the third is trained separately. The first two models are based on the ideas of CycleGAN, where network performs transferring from the domain of artificial signs to the domain of real ones. The third model is fundamentally different and inspired by the ideas of StyleGAN, in which the neural network itself learns how to generate correct traffic sign icons that are consistent with the background.

All proposed approaches take into account the context of the image around the embedded sign. That is the main difference from existing methods for artificial traffic sign processing. 

\subsubsection{First approach ("Pasted")}
In this approach, we train together neural networks for inpainting and processing of embedded traffic sign. We use Edgeconnect \cite{edgeconnect} as the basis for the inpainting architecture. We remove the part of the model for object boundaries generation. The whole proposed architecture consists of two generators and two discriminators.

The first generator receives an input image patch of size $128 \times 128$ pixels and a mask of a removed part in the middle of it. The output from the first generator inpaints the removed part. The first discriminator receives either the inpainted patch of an image or the original patch without the removed part and learns to distinguish the real ones from the generated ones. During training, this patch is cut out from random places of source pictures, and a random rectangle is removed from it exactly in the middle so that each side of the removed part is not more than $64$ pixels.

The icon of a traffic sign is then embedded in the middle of the output of the first generator, so that icon’s maximum side is $64$ pixels minus a small random number. This patch with an icon is fed to the input of the second generator, which should improve the visual quality of the fragment. The second discriminator receives at the input either output of the second generator, or a real patch with a traffic sign and learns to distinguish them from each other.

Both generators will inevitably change background with uncut part of the image, so the pixels around the cut-out patch are restored by the mask of the removed part.

We chose cross-entropy as an adversarial loss function of both discriminators. Additionally, the first generator has an $L_1-loss$, $perception\ loss$ and $style\ loss$ \cite{perceptionloss} between the inpainted by the first generator patch and the correct image. For the second generator, there is an $L_1-loss$ for the background around the sign so that it does not change. Also, a $perception\ loss$ was added between the input and output of the second generator and $style\ loss$ between the output of the second generator and the output of
the first generator, before embedding the sign icon.

\begin{figure*}[h]
\begin{center}
\includegraphics[height=6.5cm]{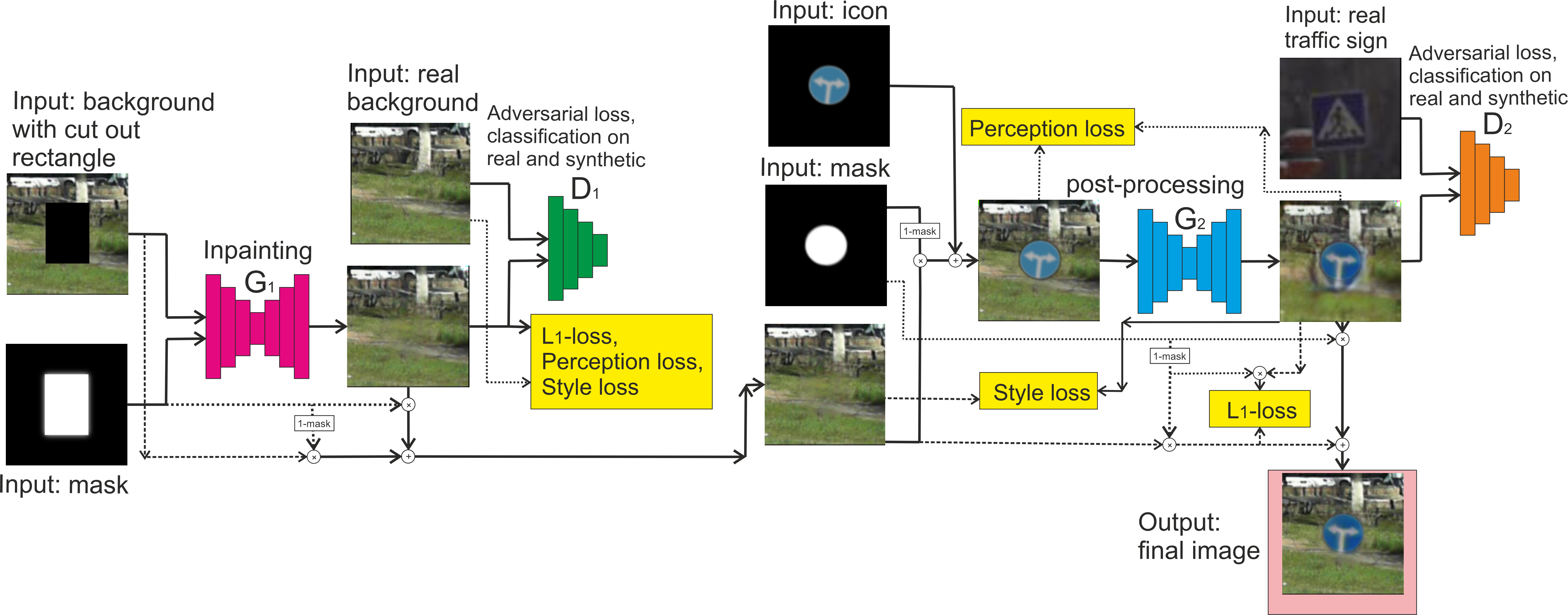}
\caption{The architecture of the first approach.}
\label{fig:met1_common}
\end{center}
\end{figure*}

The architecture diagrams of neural networks are shown in figure
\ref{fig:met1_common}. During inference and generating of the synthetic data set, patches with real traffic signs are cut out of the image and replaced with patches with embedded artificial signs.  

\subsubsection{Second approach ("Cycled")}
In the first approach, real traffic signs in patches are used only in the second discriminator. That means that the first generator performs inpainting of patches for which the correct background is known in advance, but the second generator embeds the sign when the true result of embedding is unknown. We decided to add a second data stream to the training process, where the input will be fed with the patch, in which the real sign was previously located, but was cut out. Next, for the second data stream, the inpainting of the cut-out part of the fragment of the picture is performed by the first generator. Here, unlike the first stream, the true output of the first generator is unknown. Then the icon of the sign of the same class, which was in a real patch, is embedded. This icon is processed by the second generator. As a result, the entire neural network should ideally get a picture identical to the original one.

In addition, $L_1-loss$, $perception\ loss$ and $style\ loss$ between the outputs of the second generator and the real image were added as loss functions for the second data stream. Also, we added $L_1-loss$ between the input and output of the first generator around the area of cut out a rectangle in a fragment.

The architecture of the neural network itself does not differ from the first approach. Cross-entropy is used similarly to the first as an adversarial loss function of discriminators in the second data stream.

\begin{figure*}[h]
\begin{center}
\includegraphics[height=6cm]{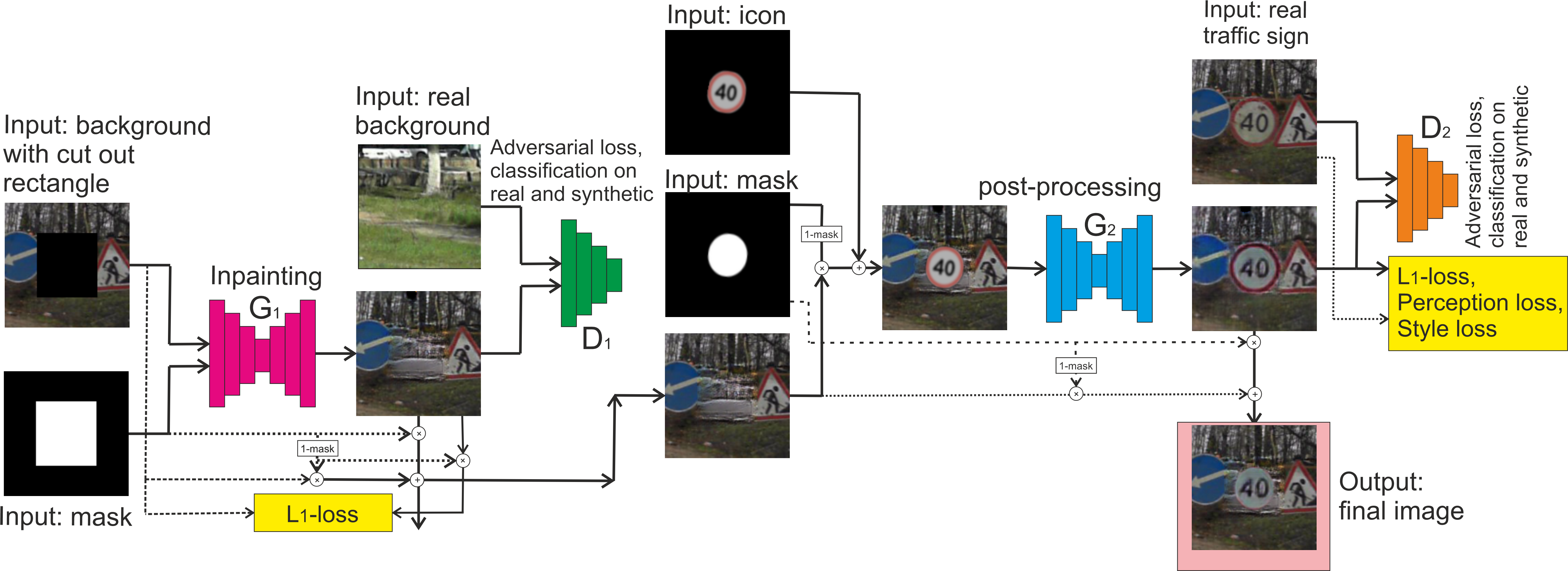}
\caption{Additional data route in second approach.}
\label{fig:met2_common}
\end{center}
\end{figure*}

The scheme of the second data stream is shown in the figure
\ref{fig:met2_common}.

\subsubsection{Third approach ("Styled")}
Two previous models have shown good quality already, but we have decided to use a more advanced generator to push the quality further. Both previous models combine two neural networks for inpainting and processing images, which are trained simultaneously. In this method, we train two parts separately. As a neural network for inpainting, we use an architecture similar to the previous approaches, based on the EdgeConnect.

Let us consider in more detail the second neural network for the processing of embedded signs in patches. We use StyleGAN \cite{stylegan} as the basis for this model. It will not process an icon which already embedded in the background, but it will generate a traffic sign consistent with the background. To achieve this result, we have made several significant changes to the StyleGAN:
\begin{itemize}
    \item Instead of generating a feature vector from random noise as in the original fully connected neural network, we propose to use two convolutional neural subnets. The first convolutional subnet gets as input an image of a $64 \times 64$-sized icon embedded in a $128 \times 128$-sized background patch, where the real traffic sign used to be located before. Subnet converts it to a vector of length $548$. The second convolutional subnet receives resized to $64 \times 64$ input fragment (the real size is $128 \times 128$) of the background without a sign.  At the output from it, a vector of length $64$ is obtained. Next, the two resulting vectors are concatenated into the vector $v_{desc}$ of length $612$.
  \item A simple two-layer classifier has been added, which, using the vector $v_{desc}$, tries to determine the class of a traffic sign from $205$ possible. This classifier improved the quality of the generated images. It seems to us that this is happening because it regularizes the neural network so that it encodes exactly the properties related to the class of the sign and not its appearance.
    \item The process of generating a sign does not begin with a trained constant activation $4 \times 4$ map, but with a map obtained from $v_{desc}$ using one fully-connected layer.
    \item An additional second discriminator is added, which distinguishes the synthetic sign embedded into the background patch from the real one.
    \item As in the original StyleGAN, all parts of the neural network are first trained for small $8 \times 8$ pictures, then the layers of generators and discriminators are gradually turned on up to the size of the $64 \times 64$ sign icon located in the center of the background with a size of $128 \times 128$.
\end{itemize}
An adversarial loss function was WGAN-GP in both discriminators. Also with $VGG13$ neural network we added $perception\ loss$ between the output of the neural network and the icon, embedded into the background without processing. Additionally, we used a small weight $perception\ loss$ between the background itself and the output of a neural network with a sign. It has been observed that this increases the realism of the generated images.

During training, synthetic traffic signs are located in places where there used to be real traffic signs. That is why we previously performed inpainting of real signs.

\begin{figure*}[h]
\begin{center}
\includegraphics[height=7.5cm]{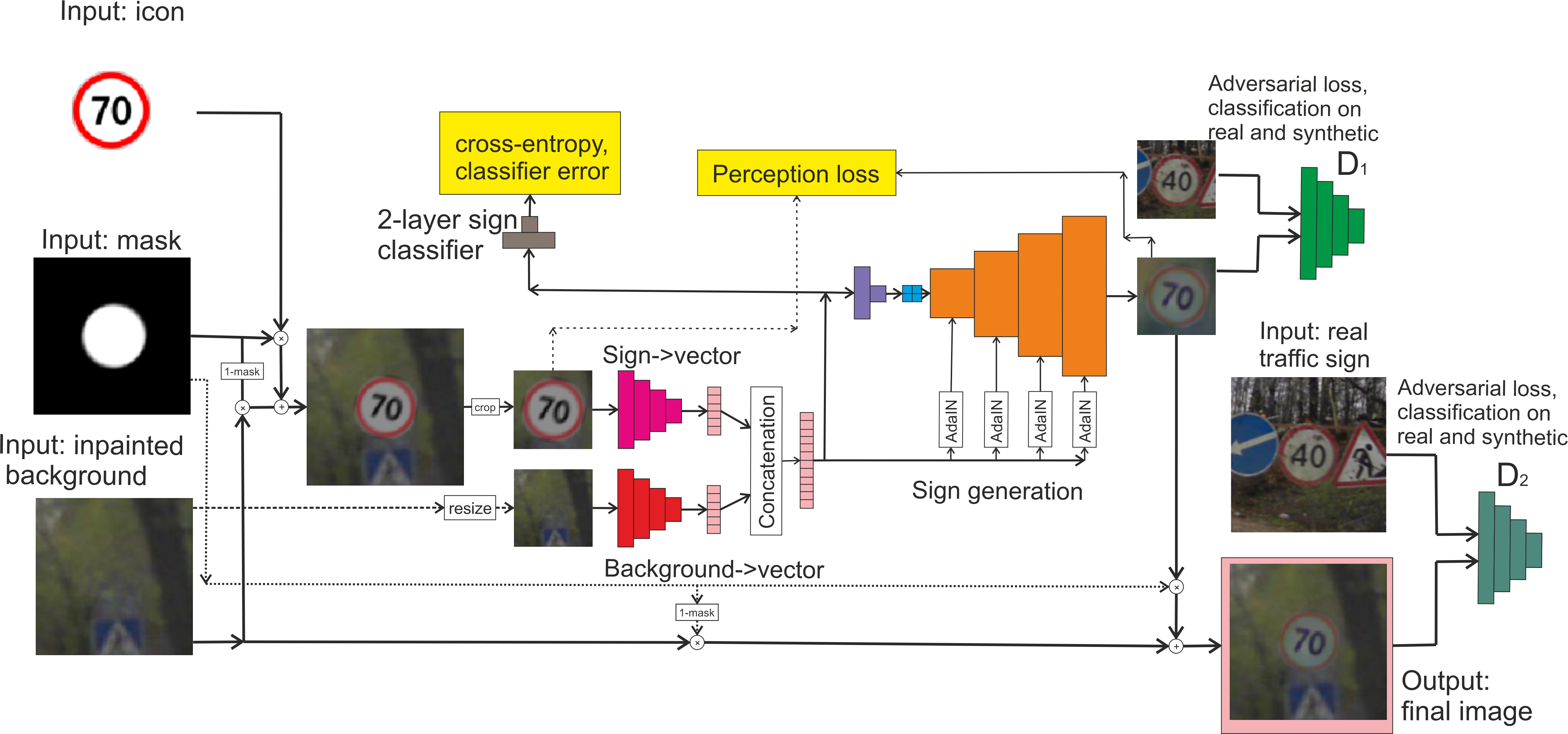}
\caption{Architecture of generator for processing in third approach.}
\label{fig:met3_common}
\end{center}
\end{figure*}

The processing scheme of traffic signs in the third model is shown in figure \ref{fig:met3_common}. The proposed method allowed a generation of images with good quality and exceeded two previous models in experimental evaluations.

\subsection{Location of embedded traffic signs in real road images}
We have also examined the geometric positioning of traffic signs in the image. We train a neural network that will find appropriate places for additional traffic signs on road images.

We have used sampling with kernel density estimation from the distribution of existing labelled data positions as a baseline for the placement of additional traffic signs. This approach does not take into account the features of each particular image. It was trained on real labelled training samples. We built three different distributions -- to sample coordinates of the signs in the image, sample their sizes and the number of signs in the current image.

Next, we have tried the approach from \cite{placementcaware}. In our work, we have used the only \emph{where} module, while \emph{what} module was disabled. We didn't find any papers, where such type of neural networks was used for traffic sign placement previously.

For the given image, the model tries to predict the correct distribution of sizes and locations of object instances. It is a generative adversarial network, where the generator as input takes semantic segmentation of image and a random vector. As output, it returns parameters of affine transformation without rotation for an appropriate bounding box for a new sign.

This architecture has two discriminators. First, $D_1$ learns to differentiate real and generated affine parameters for the current image. Second, $D_2$ learns to distinguish whether a new bounding object is consistent with the input semantic map. Cross-entropy is used as an adversarial loss for them. During training, this module has two paths -- unsupervised and supervised. An unsupervised path has only a second
discriminator $D_2$, while a supervised has both $D_1$ and $D_2$.

For unsupervised path, architecture has input reconstruction loss which aims to reconstruct input semantic map and random vector from intrinsic representation for STN subnetwork using $L_1-loss$. This helps to ensure that encoded representation has significant information from input data and partially solves the problem of model collapsing to a few numbers of modes and not covering the entire distribution.

In the supervision path, we already have one of the real positions of traffic signs. This information should be conveyed to architecture. To achieve this, the network has an additional submodule that encodes real affine transformation to the input vector (instead of random) and output transformation should be the same as the input. Kullback-Leibler divergence term in loss helps this submodule to learn the correct distribution for encoding. $D_1$ tries to distinguish synthetic parameters. This path also helps positions determined by the transform to become
more diverse.

This neural network for determining the location of objects is based on semantic maps. Since the RTSD dataset does not have semantic segmentation, we first conducted experiments in which RGB road images are fed to the input of a neural network. With such training, we were not able to achieve acceptable quality and the generated distributions themselves collapsed into degenerate when all new signs are located in the same place for every image.

To solve the problem of missing RTSD semantic segmentation, we have applied to our dataset the semantic segmentation model, trained on Cityscapes dataset. We have used the pre-trained method 'Fast Semantic Segmentation' \cite{fastsemseg}. It generates plausible semantic segmentation. We have used the obtained semantic maps to train \emph{where} module of the neural network for object placement.

\begin{figure*}[h]
\begin{center}
\includegraphics[height=12cm]{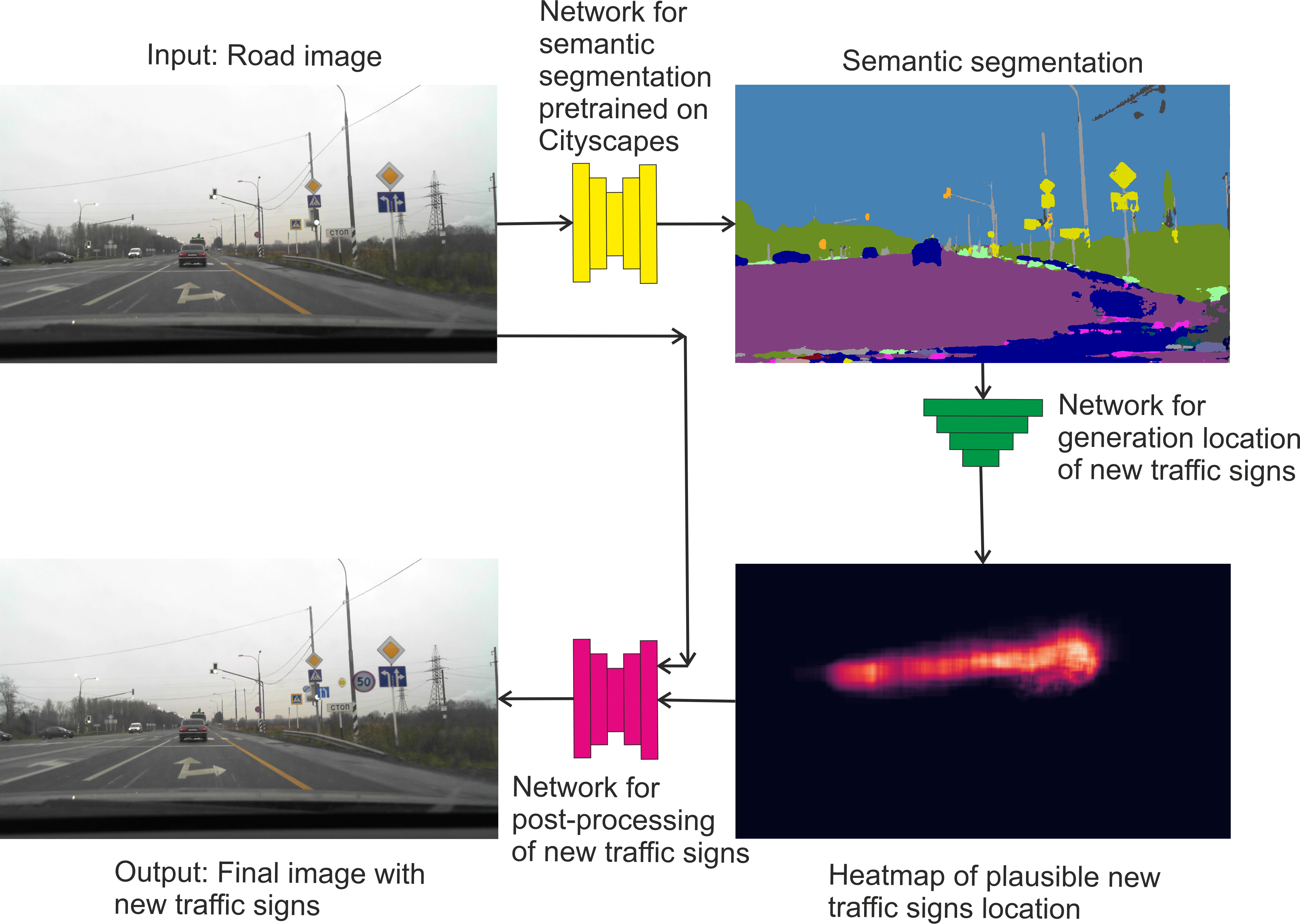}
\caption{A full pipeline of proposed traffic sign placement method.}
\label{fig:placement_dataroute}
\end{center}
\end{figure*}

After that, we have used a trained neural network to sample the locations and sizes of new traffic signs. When generating them, we have made sure that the new examples did not overlap. The number of traffic signs for each image has been  determined using Gaussian kernel density estimation. A full pipeline of new traffic sign generation process is portrayed in figure \ref{fig:placement_dataroute}.

\section{Evaluation}
\subsection{RTSD Dataset}
As real data, we took Russian traffic sign dataset RTSD \cite{rtsd}. It consists of 205 classes, of which 99 are found only in the test set and are completely absent in the training set, and 106 classes are present in the training set. The
set contains data for training detectors and classifiers of traffic signs. Train and test data statistics can be found in tables \ref{tab:datasetstatistics1},
\ref{tab:datasetstatistics2}.

\begin{table}[h]
\begin{center}
\begin{tabular}{|
>{\columncolor[HTML]{D0D0D0}}c |c|c|}
\hline
\multicolumn{1}{|l|}{\cellcolor[HTML]{D0D0D0}}                              & \cellcolor[HTML]{D0D0D0}Images & \cellcolor[HTML]{D0D0D0}Signs \\ \hline
\begin{tabular}[c]{@{}c@{}}Train set, Pasted,\\ Cycled, Styled\end{tabular} & 47639                          & 80277                         \\ \hline
Test set                                                                    & 11389                          & 25232                         \\ \hline
\end{tabular}
\end{center}
\caption{Statistics for detection task RTSD dataset.} \label{tab:datasetstatistics1}
\end{table}

\begin{table}[h]
\begin{center}
\begin{tabular}{|
>{\columncolor[HTML]{D0D0D0}}c |c|c|c|}
\hline
\multicolumn{1}{|l|}{\cellcolor[HTML]{D0D0D0}}                    & \cellcolor[HTML]{D0D0D0}All & \cellcolor[HTML]{D0D0D0}Rare & \cellcolor[HTML]{D0D0D0}Frequent \\ \hline
Train set                                                         & 79896                       & 0                            & 79896                            \\ \hline
Test set                                                          & 25613                       & 1622                         & 23991                            \\ \hline
CGI-GAN                                                          & 193444                      & 94465                        & 98979                            \\ \hline
\begin{tabular}[c]{@{}c@{}}Pasted, Cycled,\\  Styled\end{tabular} & 196455                      & 94472                        & 101983                           \\ \hline
\end{tabular}
\end{center}
\caption{Statistics for classification task RTSD dataset.} \label{tab:datasetstatistics2}
\end{table}

Also for all 205 classes, we had high-resolution icons of traffic signs with their masks.

We compared our proposed approach for embedding synthetic objects in pictures with three already existed methods for traffic sign \cite{icviprtsd} processing:
\begin{itemize}
    \item \textbf{Synt} -- this is a simple synthetic, which was obtained by
        embedding signs on the background and applying a transformation of sign
        with random parameters to the icon: rotate, shift, contrast change,
        Gaussian blur, motion blur.
    \item \textbf{CGI} -- samples, which were obtained by rendering
        three-dimensional models of traffic signs on pillars in real road
        images.
    \item \textbf{CGI-GAN} -- in this sample, traffic signs are transformed from
        the $CGI$ collection to better ones using CycleGAN.
    \item \textbf{Inpaint} -- this is a simple synthetic data for the detector, in
        which an icon of a traffic sign is drawn in the image without any
        processing.
\end{itemize}

\subsection{Generated data sets}
To begin with, we conducted experiments in which synthetic traffic signs were embedded in places where real ones already existed. This secured the correct geometric placement of synthetic signs. For detector, the number of images and signs in the synthetic set is the same as in the real dataset. For classifier, the number of samples is the same as for previously existed synthetic data sets. Let's introduce abbreviations for the proposed three models:
\begin{itemize}
    \item \textbf{Pasted} -- results of first approach.
    \item \textbf{Cycled} -- results of second approach.
    \item \textbf{Styled} -- results of third approach.
\end{itemize}

Then we generated a synthetic set for the detector in which new places of traffic signs were determined or using kernel density estimation (\textbf{KDE}) or using a special neural network (\textbf{NN}). At the same time, the processing of synthetic signs was made only by the third \textbf{Styled} approach, which
showed itself best for rare traffic signs. In this method, we conducted various experiments. Let's introduce abbreviations for proposed approaches:
\begin{itemize}
    \item \textbf{KDE-additional} and \textbf{NN-additional} -- Locate new
        traffic signs in addition to existing ones.
    \item \textbf{KDE-only-synt} and \textbf{NN-only-synt} -- Perform
        inpainting of real signs to remove them, and then place synthetic signs
        in new places. As a result, there are no signs in such pictures at the
        places of previously existed.
    \item \textbf{KDE-manystyled} and \textbf{NN-manystyled} -- Perform
        inpainting of real signs, and then place synthetic signs in both new
        places and existed in real places.
\end{itemize}

The number of images in each set was the same as the number of images in the training set, because each training image was augmented exactly one time.

\subsection{Traffic sign recognition system}
As an object detector, we use PVANet \cite{pvanet}, which is based on the Faster R-CNN approach. We evaluated detection output on a test set before and after we applied the classifier. The area under curve (AUC) was used to measure detector quality.

As traffic sign classifiers we chose two models based on WideResNet \cite{wideresnet}. The first one is a simple classifier model with WideResNet architecture. It takes an image of size $64 \times 64$ pixels and predicts one of the $205$ sign classes. On the features extracted by this neural network, we trained a simple k-NN classifier. It operates on an index that consists of synthetic examples of traffic signs. The second method is designed specifically to handle the case of rare traffic sign classes. It is proposed in paper \cite{betterwideresnet}. In this method, rare and frequent classes are treated differently. First, WideResNet features are extracted at a penultimate layer of the neural network. These features are then used in Random Forest to classify whether a sign is rare or frequent. Frequent signs are classified with the Softmax layer on top of WideResNet. Rare classes are passed into a k-NN classifier. This classifier shows better quality compared to the first classifier \cite{betterwideresnet}.
To measure quality, we first calculated overall accuracy on the test set. In the same tables, separately for rare and frequent classes, we calculated the micro-averaged Recall (formula \ref{eq:micro_recall}), as it is important for us to understand how many signs we find from the available ones. $M$ is the number of classes. For class $i$ we define $TP_i$, $FN_i$, $FP_i$ as the number of true positives, false negatives and false positives respectively.
\begin{equation} \label{eq:micro_recall}
micro-averaged~Recall = \frac{\sum_{i=1}^{M} TP_i}{\sum_{i=1}^{M} (TP_i + FN_i)}
\end{equation}
Next, we compared the macro-averaged Precision(formula \ref{eq:macro_precision}), Recall(formula \ref{eq:macro_recall}), and F1(formula \ref{eq:macro_f1}) measures for all classes and separately for rare and frequent classes.
\begin{equation} \label{eq:macro_precision}
macro-averaged~Precision = \frac{\sum_{i=1}^{M} \frac{TP_i}{TP_i+FP_i}}{M},
\end{equation}
\begin{equation} \label{eq:macro_recall}
macro-averaged~Recall = \frac{\sum_{i=1}^{M} \frac{TP_i}{TP_i+FN_i}}{M}, 
\end{equation}
\begin{equation} \label{eq:macro_f1}
macro-averaged~F1 = \frac{\sum_{i=1}^{M} \frac{2 * Precision_i * Recall_i}{Precision_i+Recall_i}}{M}, 
\end{equation}

\section{Evaluation results}
\begin{figure*}[h]
\begin{center}
\setlength\tabcolsep{0pt}
\renewcommand{\arraystretch}{0}
\begin{tabular}{
        >{\centering\arraybackslash}m{1.4cm}
        >{\centering\arraybackslash}m{1.4cm}
        >{\centering\arraybackslash}m{1.4cm}
        >{\centering\arraybackslash}m{1.4cm}
        >{\centering\arraybackslash}m{1.4cm}
        >{\centering\arraybackslash}m{1.4cm}
        >{\centering\arraybackslash}m{1.4cm}
        >{\centering\arraybackslash}m{1.4cm}
        }
\includegraphics[width=14mm]{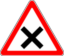} &
\includegraphics[height=14mm]{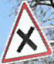} &
\includegraphics[width=14mm]{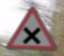} &
\includegraphics[height=14mm]{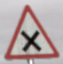} &
\includegraphics[width=14mm]{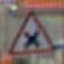} &
\includegraphics[width=14mm]{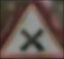} &
\includegraphics[width=14mm]{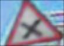} &
\includegraphics[width=14mm]{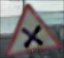} \\

\includegraphics[width=14mm]{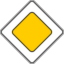} &
\includegraphics[width=14mm]{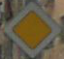} &
\includegraphics[width=14mm]{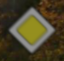} &
\includegraphics[height=14mm]{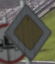} &
\includegraphics[width=14mm]{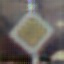} &
\includegraphics[height=14mm]{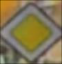} &
\includegraphics[width=14mm]{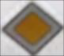} &
\includegraphics[height=14mm]{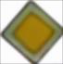} \\

\includegraphics[width=14mm]{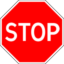} &
\includegraphics[width=14mm]{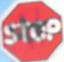} &
\includegraphics[height=14mm]{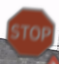} &
\includegraphics[height=14mm]{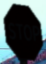} &
\includegraphics[width=14mm]{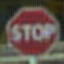} &
\includegraphics[height=14mm]{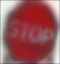} &
\includegraphics[width=14mm]{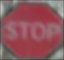} &
\includegraphics[height=14mm]{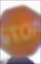} \\

\includegraphics[width=14mm]{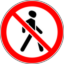} &
\includegraphics[width=14mm]{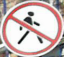} &
\includegraphics[height=14mm]{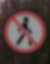} &
\includegraphics[height=14mm]{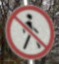} &
\includegraphics[width=14mm]{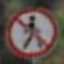} &
\includegraphics[height=14mm]{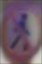} &
\includegraphics[width=14mm]{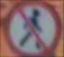} &
\includegraphics[height=14mm]{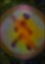} \\

\includegraphics[width=14mm]{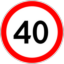} &
\includegraphics[height=14mm]{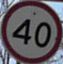} &
\includegraphics[width=14mm]{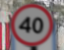} &
\includegraphics[height=14mm]{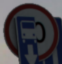} &
\includegraphics[width=14mm]{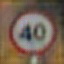} &
\includegraphics[height=14mm]{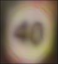} &
\includegraphics[width=14mm]{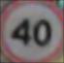} &
\includegraphics[width=14mm]{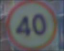} \\

\includegraphics[width=14mm]{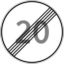} &
\includegraphics[height=14mm]{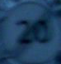} &
\includegraphics[height=14mm]{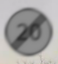} &
\includegraphics[height=14mm]{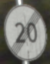} &
\includegraphics[width=14mm]{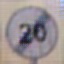} &
\includegraphics[height=14mm]{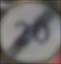} &
\includegraphics[height=14mm]{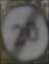} &
\includegraphics[height=14mm]{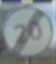} \\

\includegraphics[width=14mm]{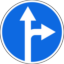} &
\includegraphics[height=14mm]{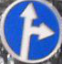} &
\includegraphics[height=14mm]{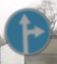} &
\includegraphics[width=14mm]{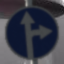} &
\includegraphics[width=14mm]{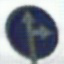} &
\includegraphics[height=14mm]{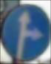} &
\includegraphics[height=14mm]{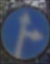} &
\includegraphics[height=14mm]{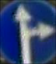} \\

\includegraphics[height=14mm]{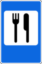} &
\includegraphics[height=14mm]{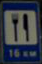} &
\includegraphics[height=14mm]{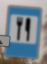} &
\includegraphics[height=14mm]{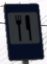} &
\includegraphics[width=14mm]{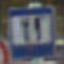} &
\includegraphics[height=14mm]{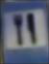} &
\includegraphics[height=14mm]{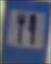} &
\includegraphics[height=14mm]{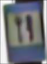} \\

\includegraphics[width=14mm]{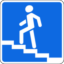} &
\includegraphics[height=14mm]{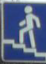} &
\includegraphics[height=14mm]{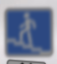} &
\includegraphics[height=14mm]{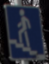} &
\includegraphics[width=14mm]{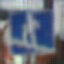} &
\includegraphics[height=14mm]{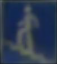} &
\includegraphics[height=14mm]{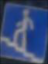} &
\includegraphics[width=14mm]{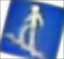} \\

\includegraphics[width=14mm]{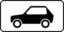} &
\includegraphics[width=14mm]{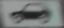} &
\includegraphics[width=14mm]{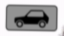} &
\includegraphics[width=14mm]{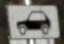} &
\includegraphics[width=14mm]{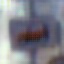} &
\includegraphics[width=14mm]{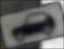} &
\includegraphics[width=14mm]{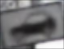} &
\includegraphics[width=14mm]{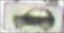} \\

icon & real & synt & CGI & CGI-GAN & Pasted & Cycled & Styled \\
\end{tabular}
\end{center}\caption{Comparison of different synthetic traffic signs types.} \label{fig:comparesigns} \end{figure*}

\begin{figure*}[h]
\begin{center}
\setlength\tabcolsep{0pt}
\renewcommand{\arraystretch}{0}
\begin{tabular}{ccc}
\includegraphics[width=50mm]{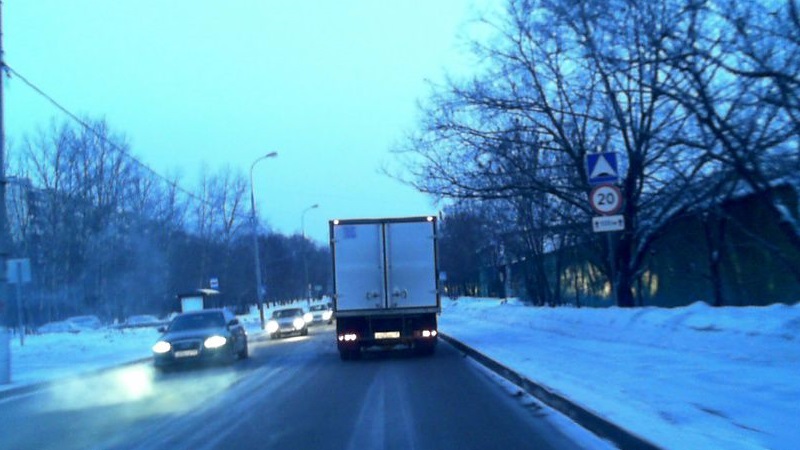} &
\includegraphics[width=50mm]{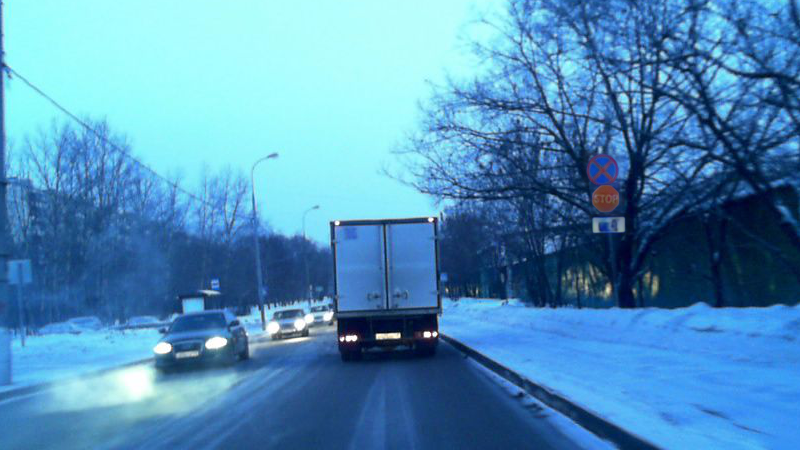} &
\includegraphics[width=50mm]{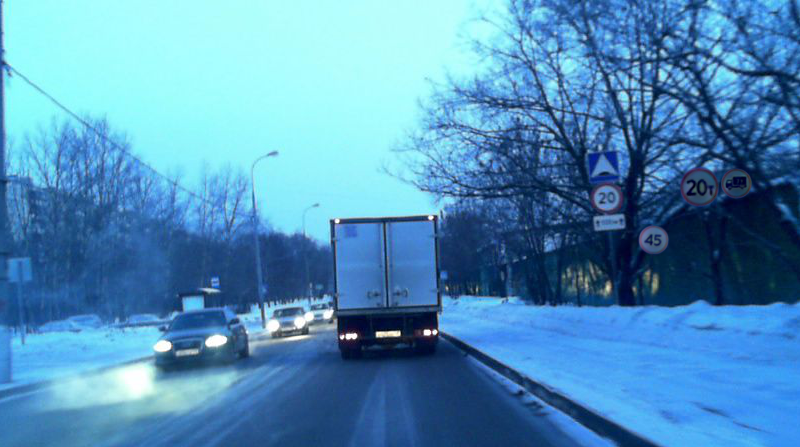} \\

\includegraphics[width=50mm]{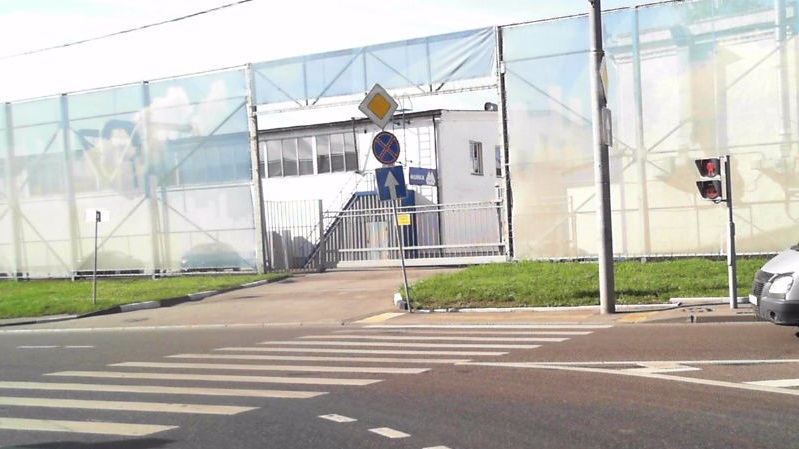} &
\includegraphics[width=50mm]{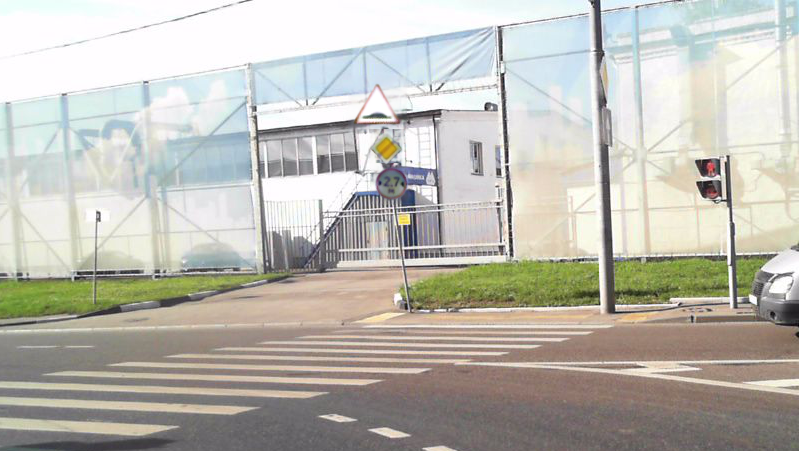} &
\includegraphics[width=50mm]{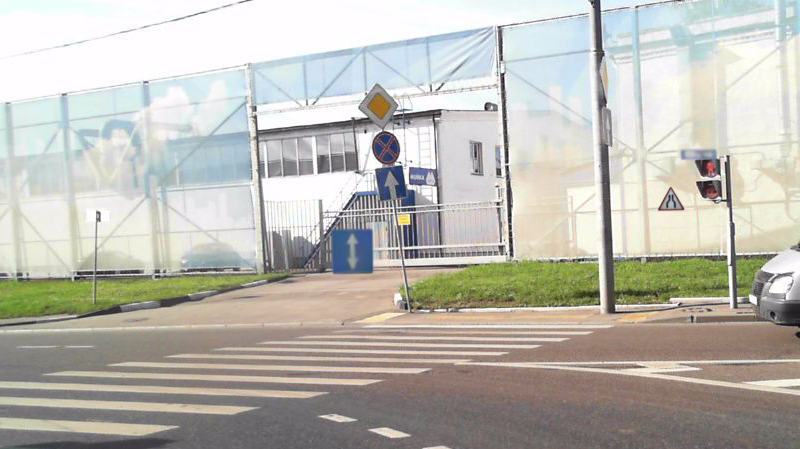} \\

\includegraphics[width=50mm]{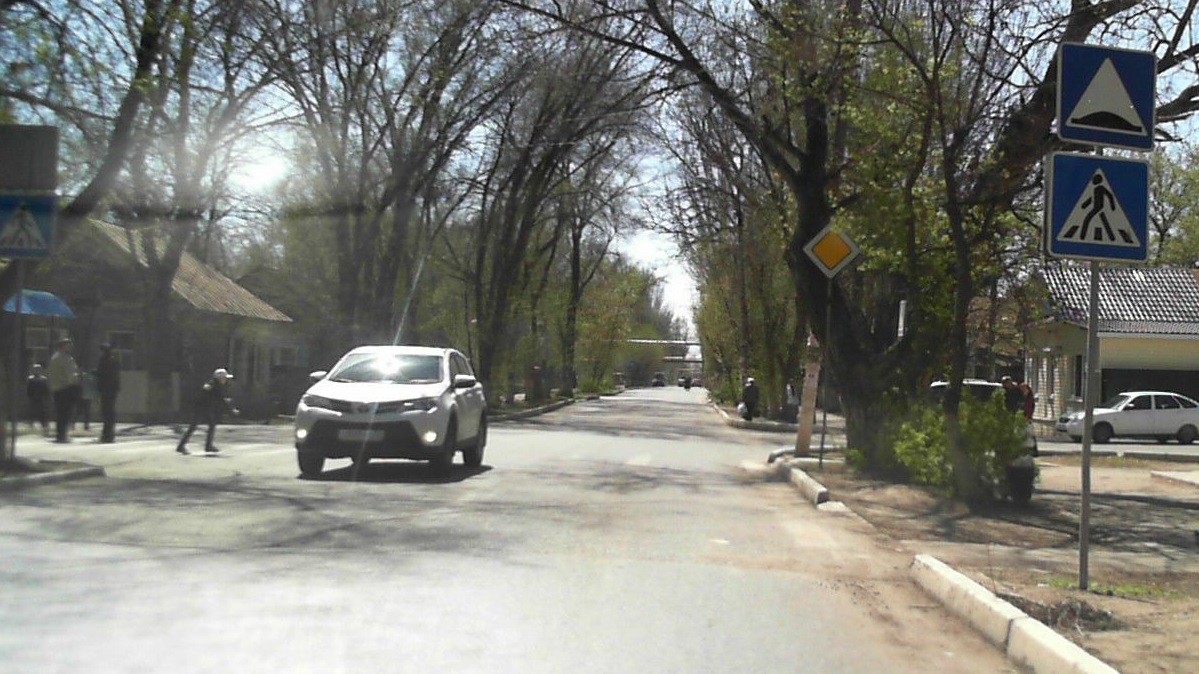} &
\includegraphics[width=50mm]{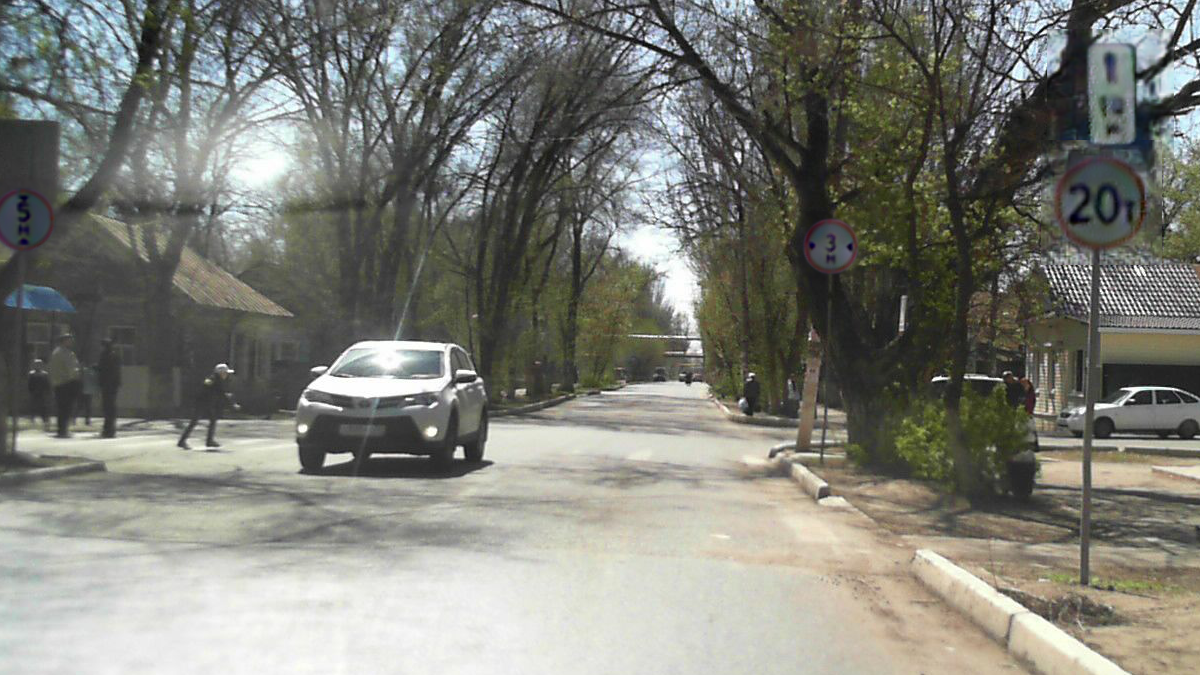} &
\includegraphics[width=50mm]{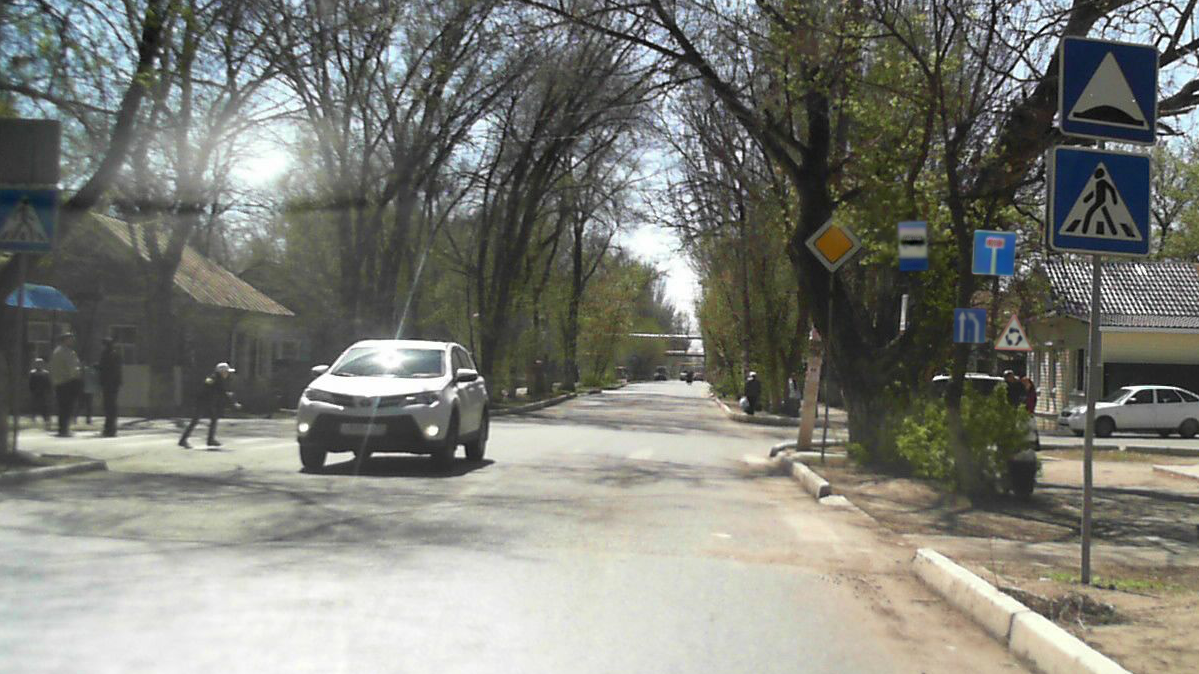} \\

\includegraphics[width=50mm]{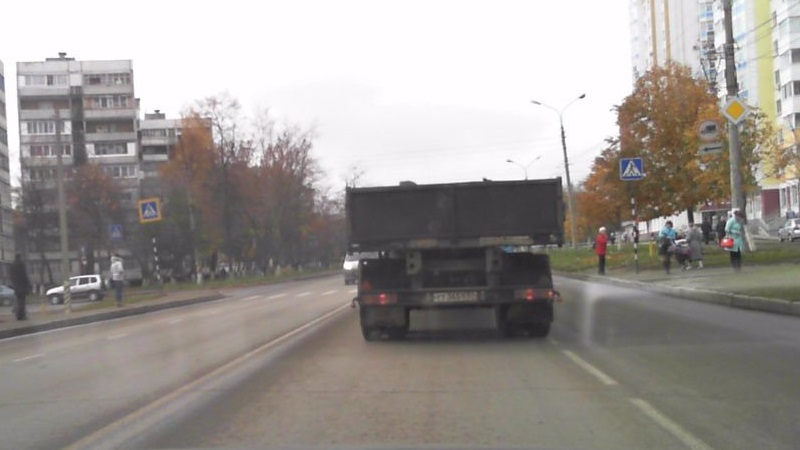} &
\includegraphics[width=50mm]{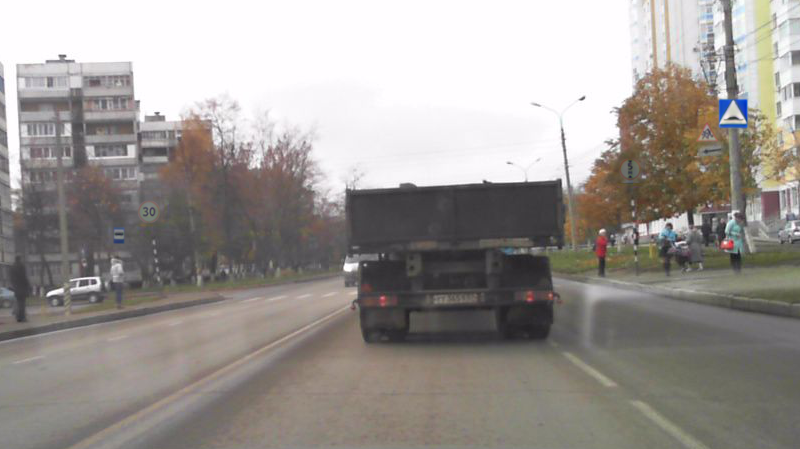} &
\includegraphics[width=50mm]{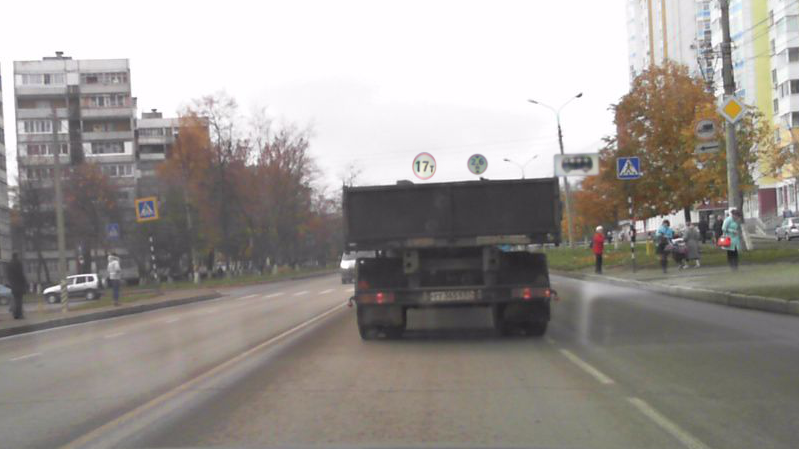} \\

\includegraphics[width=50mm]{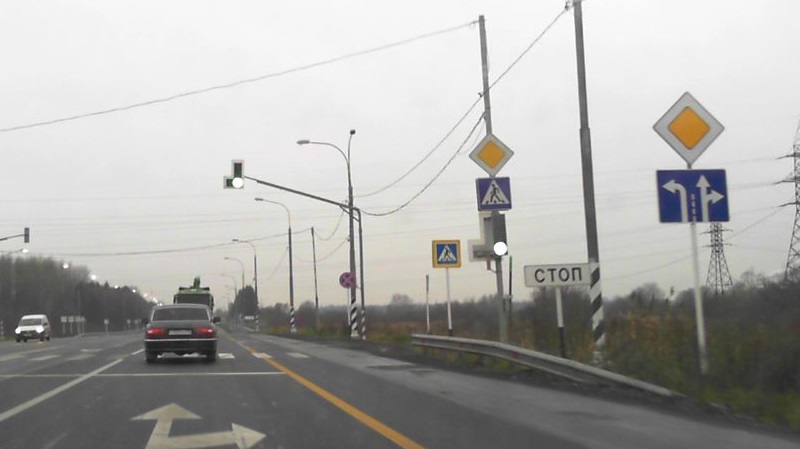} &
\includegraphics[width=50mm]{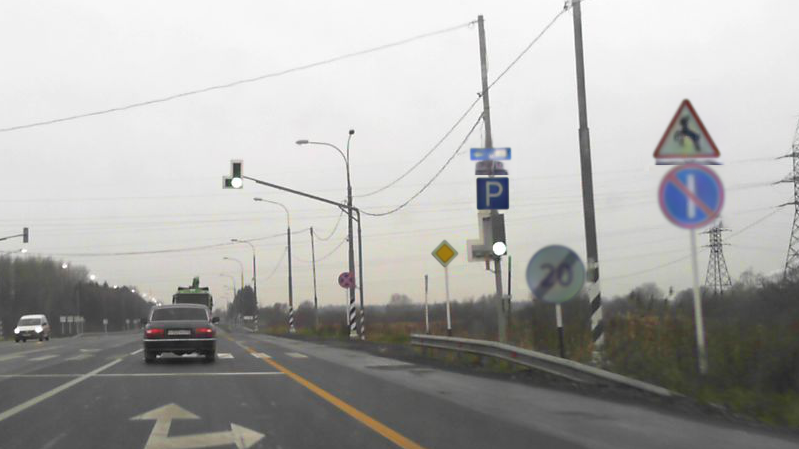} &
\includegraphics[width=50mm]{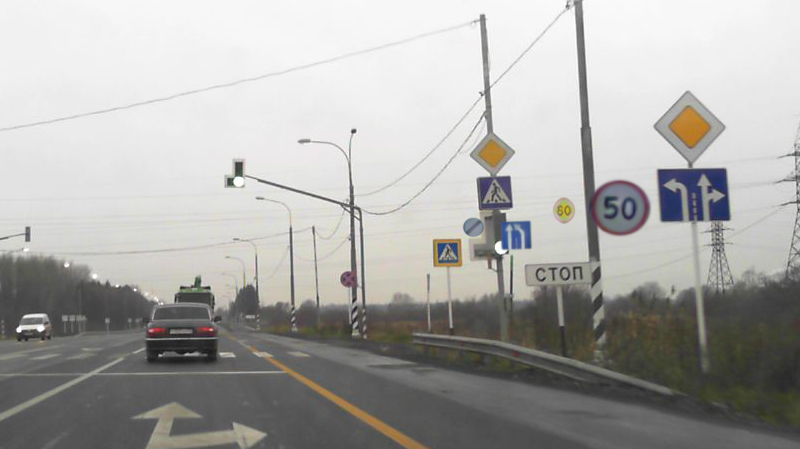} \\

real image & image with replaced sings & image with additional signs \\
\end{tabular}
\end{center}
\caption{Comparison of real images; images from \textbf{Styled} set, where real
images replaced with synthetic ones; images from \textbf{NN-additional} set,
where new traffic signs were located in addition to existing ones.}
\label{fig:compareimages}
\end{figure*}

During our experiments, we trained neural networks for classification and detection both on mixtures of real data with synthetic data and on synthetic data alone. A comparison of traffic signs examples for a classifier can be seen in figure \ref{fig:comparesigns}. Examples of road images with synthetic signs are in figure \ref{fig:compareimages}.

\subsection{Classifier results}
Here we describe the results of experiments with classifiers. We compared the proposed approach with the best previous method, which uses synthetic data $CGI-GAN$ \cite{betterwideresnet}. Table \ref{tab:simplewideresnetmix} shows measurements of a simple WideResNet classifier trained on a mixture of real and synthetic samples with a k-NN index trained on its features.
For all classes we measured accuracy, but separately for rare and frequent classes we measured micro-average recall.
Table \ref{tab:simplewideresnetonly} shows measurements of a simple WideResNet classifier, trained only on synthetic samples. Table
\ref{tab:betterwideresnetmix} summarizes measurements of improved WideResNet classifier trained on a mixture of real and synthetic samples. Table \ref{tab:betterwideresnetonly} shows metrics of an improved WideResNet classifier trained only on synthetic samples. Table \ref{tab:macrowrnetmetriconly} shows macro-averaged Precision, Recall, and F1 measures for WideResNet classifiers trained only on synthetic samples. And table \ref{tab:macrowrnetmetricmixture} for classifiers trained on a mixture of real and synthetic samples.

\begin{table*}[h]
\resizebox{\textwidth}{!}{\begin{tabular}{|
>{\columncolor[HTML]{D0D0D0}}c |c|c|c|c|c|c|c|c|c|c|c|c|}
\hline
{\color[HTML]{000000} }                & \multicolumn{3}{c|}{\cellcolor[HTML]{D0D0D0}{\color[HTML]{000000} \begin{tabular}[c]{@{}c@{}}Metrics of\\ softmax output\end{tabular}}}                                                                                                   & \multicolumn{3}{c|}{\cellcolor[HTML]{D0D0D0}{\color[HTML]{000000} \begin{tabular}[c]{@{}c@{}}Metrics with index\\ from icons\end{tabular}}}                            & \multicolumn{3}{c|}{\cellcolor[HTML]{D0D0D0}{\color[HTML]{000000} \begin{tabular}[c]{@{}c@{}}Metrics with index\\ from test set\end{tabular}}}                                        & \multicolumn{3}{c|}{\cellcolor[HTML]{D0D0D0}{\color[HTML]{000000} \begin{tabular}[c]{@{}c@{}}Metrics with index\\ from synthetic set\end{tabular}}}                                                \\ \cline{2-13}
\rowcolor[HTML]{D0D0D0}
{\color[HTML]{000000} }                & {\color[HTML]{000000} \begin{tabular}[c]{@{}c@{}}all,\\ Accuracy\end{tabular}}                                & {\color[HTML]{000000} \begin{tabular}[c]{@{}c@{}}rare,\\ Recall\end{tabular}}                                 & {\color[HTML]{000000} \begin{tabular}[c]{@{}c@{}}frequent,\\ Recall\end{tabular}}                                 & {\color[HTML]{000000} \begin{tabular}[c]{@{}c@{}}all,\\ Accuracy\end{tabular}}         & {\color[HTML]{000000} \begin{tabular}[c]{@{}c@{}}rare,\\ Recall\end{tabular}}                                 & {\color[HTML]{000000} \begin{tabular}[c]{@{}c@{}}frequent,\\ Recall\end{tabular}}                                 & {\color[HTML]{000000} \begin{tabular}[c]{@{}c@{}}all,\\ Accuracy\end{tabular}}                                & {\color[HTML]{000000} \begin{tabular}[c]{@{}c@{}}rare,\\ Recall\end{tabular}}                                 & {\color[HTML]{000000} \begin{tabular}[c]{@{}c@{}}frequent,\\ Recall\end{tabular}}                                & {\color[HTML]{000000} \begin{tabular}[c]{@{}c@{}}all,\\ Accuracy\end{tabular}}                                & {\color[HTML]{000000} \begin{tabular}[c]{@{}c@{}}rare,\\ Recall\end{tabular}}                                 & {\color[HTML]{000000} \begin{tabular}[c]{@{}c@{}}frequent,\\ Recall\end{tabular}}                                 \\ \hline
{\color[HTML]{000000} RTSD}            & {\color[HTML]{000000} 88.87}                                  & {\color[HTML]{000000} 0.00}                                   & {\color[HTML]{000000} 94.88}                                  & {\color[HTML]{000000} 71.53}           & {\color[HTML]{000000} 41.68}                                  & {\color[HTML]{000000} 73.55}                                  & {\color[HTML]{000000} 73.00}                                     & {\color[HTML]{000000} 61.13}                                  & {\color[HTML]{000000} 73.75}                                 & {\color[HTML]{000000} }                                   & {\color[HTML]{000000} }                                   & {\color[HTML]{000000} }                                   \\ \hline
{\color[HTML]{000000} RTSD + CGI-GAN} & {\color[HTML]{000000} 92.75}                                  & {\color[HTML]{000000} 53.82}                                  & \cellcolor[HTML]{DAE8FC}{\color[HTML]{000000} \textbf{95.39}}                                  & {\color[HTML]{000000} 77.03}           & {\color[HTML]{000000} 61.34}                                  & {\color[HTML]{000000} 78.09}                                  & {\color[HTML]{000000} 74.83}                                  & {\color[HTML]{000000} 65.79}                                  & {\color[HTML]{000000} 75.41}                                 & {\color[HTML]{000000} 82.15}                                  & {\color[HTML]{000000} 59.62}                                  & {\color[HTML]{000000} 83.67}                                  \\ \hline
{\color[HTML]{000000} RTSD + Pasted}   & {\color[HTML]{000000} 91.67}                                  & {\color[HTML]{000000} 68.74}                                  & {\color[HTML]{000000} 93.22}                                  & {\color[HTML]{000000} 76.50}            & {\color[HTML]{000000} 68.74}                                  & {\color[HTML]{000000} 77.02}                                  & {\color[HTML]{000000} 76.34}                                  & \cellcolor[HTML]{DAE8FC}{\color[HTML]{000000} \textbf{73.34}} & {\color[HTML]{000000} 76.53}                                 & {\color[HTML]{000000} 85.42}                                  & {\color[HTML]{000000} 71.33}                                  & {\color[HTML]{000000} 86.37}                                  \\ \hline
{\color[HTML]{000000} RTSD + Cycled}   & {\color[HTML]{000000} 92.03}                                  & {\color[HTML]{000000} 68.19}                                  & {\color[HTML]{000000} 93.64}                                  & \cellcolor[HTML]{DAE8FC}\textbf{78.59} & \cellcolor[HTML]{DAE8FC}{\color[HTML]{000000} \textbf{72.19}} & \cellcolor[HTML]{DAE8FC}{\color[HTML]{000000} \textbf{79.03}} & \cellcolor[HTML]{DAE8FC}{\color[HTML]{000000} \textbf{78.12}} & {\color[HTML]{000000} 70.65}                                  & \cellcolor[HTML]{DAE8FC}{\color[HTML]{000000} \textbf{78.6}} & \cellcolor[HTML]{DAE8FC}{\color[HTML]{000000} \textbf{86.71}} & \cellcolor[HTML]{DAE8FC}{\color[HTML]{000000} \textbf{73.43}} & \cellcolor[HTML]{DAE8FC}{\color[HTML]{000000} \textbf{87.61}} \\ \hline
{\color[HTML]{000000} RTSD + Styled} & \cellcolor[HTML]{DAE8FC}{\color[HTML]{000000} \textbf{92.82}} & \cellcolor[HTML]{DAE8FC}{\color[HTML]{000000} \textbf{69.67}} & 94.39 & {\color[HTML]{000000} 76.19}           & {\color[HTML]{000000} 69.05}                                  & {\color[HTML]{000000} 76.67}                                  & {\color[HTML]{000000} 77.80}                                   & {\color[HTML]{000000} 69.27}                                  & {\color[HTML]{000000} 78.35}                                 & {\color[HTML]{000000} 85.96}                                  & {\color[HTML]{000000} 70.41}                                  & {\color[HTML]{000000} 87.01}                                  \\ \hline
\end{tabular}
}\caption{Simple WideResNet classifier trained on a mixture of real and synthetic samples with a k-NN index on its features.} \label{tab:simplewideresnetmix}
\end{table*}

\begin{table*}[h]
\resizebox{\textwidth}{!}{\begin{tabular}{|c|c|c|c|c|c|c|c|c|c|c|c|c|}
\hline
\rowcolor[HTML]{D0D0D0}
{\color[HTML]{000000} }                                      & \multicolumn{3}{c|}{\cellcolor[HTML]{D0D0D0}{\color[HTML]{000000} \begin{tabular}[c]{@{}c@{}}Metrics of\\ softmax output\end{tabular}}}                                           & \multicolumn{3}{c|}{\cellcolor[HTML]{D0D0D0}{\color[HTML]{000000} \begin{tabular}[c]{@{}c@{}}Metrics with index\\ from icons\end{tabular}}} & \multicolumn{3}{c|}{\cellcolor[HTML]{D0D0D0}{\color[HTML]{000000} \begin{tabular}[c]{@{}c@{}}Metrics with index\\ from test set\end{tabular}}}                                         & \multicolumn{3}{c|}{\cellcolor[HTML]{D0D0D0}{\color[HTML]{000000} \begin{tabular}[c]{@{}c@{}}Metrics with index\\ from synthetic set\end{tabular}}} \\ \cline{2-13}
\rowcolor[HTML]{D0D0D0}
\cellcolor[HTML]{D0D0D0}{\color[HTML]{000000} }              & {\color[HTML]{000000} \begin{tabular}[c]{@{}c@{}}all,\\ Accuracy\end{tabular}}        & {\color[HTML]{000000} \begin{tabular}[c]{@{}c@{}}rare,\\ Recall\end{tabular}}                                 & {\color[HTML]{000000} \begin{tabular}[c]{@{}c@{}}frequent,\\ Recall\end{tabular}} & {\color[HTML]{000000} \begin{tabular}[c]{@{}c@{}}all,\\ Accuracy\end{tabular}}           & {\color[HTML]{000000} \begin{tabular}[c]{@{}c@{}}rare,\\ Recall\end{tabular}}                   & {\color[HTML]{000000} \begin{tabular}[c]{@{}c@{}}frequent,\\ Recall\end{tabular}}                  & {\color[HTML]{000000} \begin{tabular}[c]{@{}c@{}}all,\\ Accuracy\end{tabular}}                                & {\color[HTML]{000000} \begin{tabular}[c]{@{}c@{}}rare,\\ Recall\end{tabular}}                                 & {\color[HTML]{000000} \begin{tabular}[c]{@{}c@{}}frequent,\\ Recall\end{tabular}}                                 & {\color[HTML]{000000} \begin{tabular}[c]{@{}c@{}}all,\\ Accuracy\end{tabular}}         & {\color[HTML]{000000} \begin{tabular}[c]{@{}c@{}}rare,\\ Recall\end{tabular}}                                 & {\color[HTML]{000000} \begin{tabular}[c]{@{}c@{}}frequent,\\ Recall\end{tabular}}         \\ \hline
\cellcolor[HTML]{D0D0D0}{\color[HTML]{000000} only CGI-GAN} & {\color[HTML]{000000} 49.80}           & {\color[HTML]{000000} 43.90}                                   & {\color[HTML]{000000} 50.19}  & {\color[HTML]{000000} 22.88}             & {\color[HTML]{000000} 23.30}                     & {\color[HTML]{000000} 22.85}                   & {\color[HTML]{000000} 44.10}                                   & {\color[HTML]{000000} 46.09}                                  & {\color[HTML]{000000} 43.97}                                  & {\color[HTML]{000000} 39.35}           & {\color[HTML]{000000} 34.09}                                  & {\color[HTML]{000000} 39.70}           \\ \hline
\cellcolor[HTML]{D0D0D0}{\color[HTML]{000000} only Pasted}   & {\color[HTML]{000000} 67.50}           & {\color[HTML]{000000} 69.05}                                  & {\color[HTML]{000000} 67.39}  & {\color[HTML]{000000} 49.87}             & {\color[HTML]{000000} 47.78}                    & {\color[HTML]{000000} 50.01}                   & {\color[HTML]{000000} 58.50}                                   & {\color[HTML]{000000} 60.93}                                  & {\color[HTML]{000000} 58.34}                                  & {\color[HTML]{000000} 66.47}           & {\color[HTML]{000000} 65.66}                                  & {\color[HTML]{000000} 66.52}          \\ \hline
\rowcolor[HTML]{DAE8FC}
\cellcolor[HTML]{D0D0D0}{\color[HTML]{000000} only Cycled}   & {\color[HTML]{000000} \textbf{73.61}} & \cellcolor[HTML]{FFFFFF}{\color[HTML]{000000} 65.60}           & \textbf{74.15}                & \textbf{64.05}                           & {\color[HTML]{000000} \textbf{54.69}}           & {\color[HTML]{000000} \textbf{64.69}}          & \cellcolor[HTML]{FFFFFF}{\color[HTML]{000000} 60.01}          & \cellcolor[HTML]{FFFFFF}{\color[HTML]{000000} 62.97}          & \cellcolor[HTML]{FFFFFF}{\color[HTML]{000000} 59.82}          & {\color[HTML]{000000} \textbf{72.65}}  & \cellcolor[HTML]{FFFFFF}{\color[HTML]{000000} 63.69}          & {\color[HTML]{000000} \textbf{73.26}} \\ \hline
\rowcolor[HTML]{FFFFFF}
\cellcolor[HTML]{D0D0D0}{\color[HTML]{000000} only Styled} & {\color[HTML]{000000} 69.94}          & \cellcolor[HTML]{DAE8FC}{\color[HTML]{000000} \textbf{69.42}} & {\color[HTML]{000000} 69.97}  & {\color[HTML]{000000} 42.68}             & {\color[HTML]{000000} 48.27}                    & {\color[HTML]{000000} 42.30}                    & \cellcolor[HTML]{DAE8FC}{\color[HTML]{000000} \textbf{65.72}} & \cellcolor[HTML]{DAE8FC}{\color[HTML]{000000} \textbf{69.34}} & \cellcolor[HTML]{DAE8FC}{\color[HTML]{000000} \textbf{65.49}} & {\color[HTML]{000000} 71.77}           & \cellcolor[HTML]{DAE8FC}{\color[HTML]{000000} \textbf{67.39}} & {\color[HTML]{000000} 72.06}          \\ \hline
\end{tabular}
}\caption{Simple WideResNet classifier trained only on synthetic samples.} \label{tab:simplewideresnetonly}
\end{table*}

\begin{table*}[h]
\resizebox{\textwidth}{!}{\begin{tabular}{|c|c|c|c|c|c|c|c|c|c|}
\hline
\rowcolor[HTML]{D0D0D0}
{\color[HTML]{000000} }                                        & \multicolumn{3}{c|}{\cellcolor[HTML]{D0D0D0}{\color[HTML]{000000} \begin{tabular}[c]{@{}c@{}}Metrics with index\\ from icons\end{tabular}}}                                                  & \multicolumn{3}{c|}{\cellcolor[HTML]{D0D0D0}{\color[HTML]{000000} \begin{tabular}[c]{@{}c@{}}Metrics with index\\ from test set\end{tabular}}}                                         & \multicolumn{3}{c|}{\cellcolor[HTML]{D0D0D0}{\color[HTML]{000000} \begin{tabular}[c]{@{}c@{}}Metrics with index\\ from synthetic set\end{tabular}}}                                                \\ \cline{2-10}
\rowcolor[HTML]{D0D0D0}
{\color[HTML]{000000} }                                        & {\color[HTML]{000000} \begin{tabular}[c]{@{}c@{}}all,\\ Accuracy\end{tabular}}                                & {\color[HTML]{000000} \begin{tabular}[c]{@{}c@{}}rare,\\ Recall\end{tabular}}                                & {\color[HTML]{000000} \begin{tabular}[c]{@{}c@{}}frequent,\\ Recall\end{tabular}}                                 & {\color[HTML]{000000} \begin{tabular}[c]{@{}c@{}}all,\\ Accuracy\end{tabular}}                                & {\color[HTML]{000000} \begin{tabular}[c]{@{}c@{}}rare,\\ Recall\end{tabular}}                                 & {\color[HTML]{000000} \begin{tabular}[c]{@{}c@{}}frequent,\\ Recall\end{tabular}}                                 & {\color[HTML]{000000} \begin{tabular}[c]{@{}c@{}}all,\\ Accuracy\end{tabular}}                                & {\color[HTML]{000000} \begin{tabular}[c]{@{}c@{}}rare,\\ Recall\end{tabular}}                                 & {\color[HTML]{000000} \begin{tabular}[c]{@{}c@{}}frequent,\\ Recall\end{tabular}}                                 \\ \hline
\cellcolor[HTML]{D0D0D0}{\color[HTML]{000000} RTSD + CGI-GAN} & {\color[HTML]{000000} 93.12}                                  & {\color[HTML]{000000} 70.65}                                 & \cellcolor[HTML]{FFFFFF}{\color[HTML]{000000} 94.43}          & {\color[HTML]{000000} 92.11}                                  & {\color[HTML]{000000} 76.69}                                  & {\color[HTML]{000000} 93.10}                                   & {\color[HTML]{000000} 93.52}                                  & {\color[HTML]{000000} 70.16}                                  & {\color[HTML]{000000} 95.09}                                  \\ \hline
\cellcolor[HTML]{D0D0D0}{\color[HTML]{000000} RTSD + Pasted}   & {\color[HTML]{000000} 92.75}                                  & {\color[HTML]{000000} 73.80}                                  & {\color[HTML]{000000} 94.03}                                  & {\color[HTML]{000000} 92.86}                                  & \cellcolor[HTML]{DAE8FC}{\color[HTML]{000000} \textbf{81.42}} & {\color[HTML]{000000} 93.59}                                  & {\color[HTML]{000000} 93.84}                                  & {\color[HTML]{000000} 74.97}                                  & {\color[HTML]{000000} 95.11}                                  \\ \hline
\cellcolor[HTML]{D0D0D0}{\color[HTML]{000000} RTSD + Cycled}   & \cellcolor[HTML]{DAE8FC}{\color[HTML]{000000} \textbf{93.31}} & \cellcolor[HTML]{DAE8FC}{\color[HTML]{000000} \textbf{76.70}} & \cellcolor[HTML]{DAE8FC}{\color[HTML]{000000} \textbf{94.44}} & \cellcolor[HTML]{DAE8FC}{\color[HTML]{000000} \textbf{93.24}} & {\color[HTML]{000000} 79.65}                                  & \cellcolor[HTML]{DAE8FC}{\color[HTML]{000000} \textbf{94.11}} & {\color[HTML]{000000} 93.98}                                  & {\color[HTML]{000000} 75.46}                                  & {\color[HTML]{000000} 95.23} \\ \hline
\cellcolor[HTML]{D0D0D0}{\color[HTML]{000000} RTSD + Styled} & {\color[HTML]{000000} 92.16}                                  & {\color[HTML]{000000} 75.83}                                 & {\color[HTML]{000000} 93.26}                                  & {\color[HTML]{000000} 93.04}                                  & {\color[HTML]{000000} 78.92}                                  & {\color[HTML]{000000} 93.94}                                  & \cellcolor[HTML]{DAE8FC}{\color[HTML]{000000} \textbf{94.11}} & \cellcolor[HTML]{DAE8FC}{\color[HTML]{000000} \textbf{76.33}} & \cellcolor[HTML]{DAE8FC}{\color[HTML]{000000} \textbf{95.31}}                                  \\ \hline
\end{tabular}
}\caption{Improved WideResNet classifier trained on a mixture of real and synthetic samples.} \label{tab:betterwideresnetmix}
\end{table*}

\begin{table*}[h]
\resizebox{\textwidth}{!}{\begin{tabular}{|
>{\columncolor[HTML]{D0D0D0}}c |c|c|c|c|c|c|c|c|c|}
\hline
{\color[HTML]{000000} }              & \multicolumn{3}{c|}{\cellcolor[HTML]{D0D0D0}{\color[HTML]{000000} \begin{tabular}[c]{@{}c@{}}Metrics with index\\ from icons\end{tabular}}} & \multicolumn{3}{c|}{\cellcolor[HTML]{D0D0D0}{\color[HTML]{000000} \begin{tabular}[c]{@{}c@{}}Metrics with index\\ from test set\end{tabular}}} & \multicolumn{3}{c|}{\cellcolor[HTML]{D0D0D0}{\color[HTML]{000000} \begin{tabular}[c]{@{}c@{}}Metrics with index\\ from synthetic set\end{tabular}}} \\ \cline{2-10}
\rowcolor[HTML]{D0D0D0}
{\color[HTML]{000000} }              & {\color[HTML]{000000} \begin{tabular}[c]{@{}c@{}}all,\\ Accuracy\end{tabular}}                                & {\color[HTML]{000000} \begin{tabular}[c]{@{}c@{}}rare,\\ Recall\end{tabular}}                                & {\color[HTML]{000000} \begin{tabular}[c]{@{}c@{}}frequent,\\ Recall\end{tabular}}                                 & {\color[HTML]{000000} \begin{tabular}[c]{@{}c@{}}all,\\ Accuracy\end{tabular}}                                & {\color[HTML]{000000} \begin{tabular}[c]{@{}c@{}}rare,\\ Recall\end{tabular}}                                 & {\color[HTML]{000000} \begin{tabular}[c]{@{}c@{}}frequent,\\ Recall\end{tabular}}                                 & {\color[HTML]{000000} \begin{tabular}[c]{@{}c@{}}all,\\ Accuracy\end{tabular}}                                & {\color[HTML]{000000} \begin{tabular}[c]{@{}c@{}}rare,\\ Recall\end{tabular}}                                 & {\color[HTML]{000000} \begin{tabular}[c]{@{}c@{}}frequent,\\ Recall\end{tabular}}                 \\ \hline
{\color[HTML]{000000} only cgi\_gan} & 59.68                                         & 57.77                                        & 59.81                                        & 59.58                                            & 66.84                                            & 59.12                                           & 60.55                                          & 54.69                                         & 60.94                                         \\ \hline
{\color[HTML]{000000} only pasted}   & 71.51                                         & \cellcolor[HTML]{DAE8FC}\textbf{73.06}       & 71.40                                         & 70.77                                            & \cellcolor[HTML]{DAE8FC}\textbf{78.00}           & 70.31                                           & 72.57                                          & 72.13                                         & 72.60                                          \\ \hline
{\color[HTML]{000000} only cycled}   & \cellcolor[HTML]{DAE8FC}\textbf{72.41}        & 70.59                                        & \cellcolor[HTML]{DAE8FC}\textbf{72.54}       & 72.18                                            & 77.28                                            & 71.84                                           & 72.35                                          & 71.89                                         & 72.38                                         \\ \hline
{\color[HTML]{000000} only styled} & 71.82                                         & 71.15                                        & 71.86                                        & \cellcolor[HTML]{DAE8FC}\textbf{73.84}           & 77.54                                            & \cellcolor[HTML]{DAE8FC}\textbf{73.60}           & \cellcolor[HTML]{DAE8FC}\textbf{73.03}         & \cellcolor[HTML]{DAE8FC}\textbf{73.98}        & \cellcolor[HTML]{DAE8FC}\textbf{72.96}        \\ \hline
\end{tabular}
}\caption{Improved WideResNet classifier trained only on synthetic samples.} \label{tab:betterwideresnetonly}
\end{table*}

\begin{table*}[h]
\resizebox{\textwidth}{!}{\begin{tabular}{|c|c|c|c|c|c|c|c|c|c|c|c|c|c|c|c|c|c|c|}
\hline
\rowcolor[HTML]{D0D0D0} 
\cellcolor[HTML]{D0D0D0}                   & \multicolumn{9}{c|}{\cellcolor[HTML]{D0D0D0}Metrics of softmax output}                                                                                                                                                                                                                                                                                                        & \multicolumn{9}{c|}{\cellcolor[HTML]{D0D0D0}Metrics with index from synthetic set}                                                                                                                                                                                                                                                                                             \\ \cline{2-19} 
\rowcolor[HTML]{D0D0D0} 
\cellcolor[HTML]{D0D0D0}                   & \multicolumn{3}{c|}{\cellcolor[HTML]{D0D0D0}all}                                                                         & \multicolumn{3}{c|}{\cellcolor[HTML]{D0D0D0}rare}                                                                       & \multicolumn{3}{c|}{\cellcolor[HTML]{D0D0D0}frequent}                                                                    & \multicolumn{3}{c|}{\cellcolor[HTML]{D0D0D0}all}                                                                         & \multicolumn{3}{c|}{\cellcolor[HTML]{D0D0D0}rare}                                                                        & \multicolumn{3}{c|}{\cellcolor[HTML]{D0D0D0}frequent}                                                                    \\ \cline{2-19} 
\rowcolor[HTML]{D0D0D0} 
\multirow{-3}{*}{\cellcolor[HTML]{D0D0D0}} & precision                              & recall                                 & F1                                     & precision                              & recall                                 & F1                                    & precision                              & recall                                 & F1                                     & precision                              & recall                                 & F1                                     & precision                              & recall                                 & F1                                     & precision                              & recall                                 & F1                                     \\ \hline
\rowcolor[HTML]{D0D0D0} 
\multicolumn{19}{|c|}{\cellcolor[HTML]{D0D0D0}Simple WideResNet classifier}                                                                                                                                                                                                                                                                                                                                                                                                                                                                                                                                                                                                                                                                                                                                 \\ \hline
\cellcolor[HTML]{D0D0D0}RTSD + CGI-GAN     & 69.98                                  & 70.27                                  & 67.80                                  & 51.11                                  & 52.06                                  & 47.80                                 & 87.60                                  & 87.28                                  & 86.48                                  & 60.94                                  & 69.03                                  & 60.68                                  & 41.22                                  & 57.29                                  & 42.05                                  & 79.36                                  & 80.00                                  & 78.07                                  \\ \hline
\cellcolor[HTML]{D0D0D0}RTSD + Pasted      & 72.79                                  & 75.88                                  & 71.83                                  & 53.95                                  & 60.06                                  & 52.22                                 & 90.39                                  & 90.65                                  & 90.14                                  & 62.22                                  & 75.92                                  & 65.09                                  & 40.02                                  & 64.24                                  & 44.94                                  & 82.95                                  & 86.83                                  & 83.91                                  \\ \hline
\cellcolor[HTML]{D0D0D0}RTSD + Cycled      & 74.01                                  & 75.12                                  & 72.01                                  & 56.78                                  & 58.32                                  & 52.76                                 & 90.11                                  & 90.80                                  & 89.99                                  & \cellcolor[HTML]{DAE8FC}\textbf{64.38} & 74.44                                  & 66.11                                  & \cellcolor[HTML]{DAE8FC}\textbf{43.28} & 61.22                                  & \cellcolor[HTML]{DAE8FC}\textbf{46.48} & 84.09                                  & 86.79                                  & 84.44                                  \\ \hline
\cellcolor[HTML]{D0D0D0}RTSD + Styled      & \cellcolor[HTML]{DAE8FC}\textbf{76.11} & \cellcolor[HTML]{DAE8FC}\textbf{76.67} & \cellcolor[HTML]{DAE8FC}\textbf{74.11} & \cellcolor[HTML]{DAE8FC}\textbf{59.45} & \cellcolor[HTML]{DAE8FC}\textbf{60.31} & \cellcolor[HTML]{DAE8FC}\textbf{55.60} & \cellcolor[HTML]{DAE8FC}\textbf{91.67} & \cellcolor[HTML]{DAE8FC}\textbf{91.95} & \cellcolor[HTML]{DAE8FC}\textbf{91.38} & 63.65                                  & \cellcolor[HTML]{DAE8FC}\textbf{76.56} & \cellcolor[HTML]{DAE8FC}\textbf{66.40} & 40.65                                  & \cellcolor[HTML]{DAE8FC}\textbf{64.46} & 45.84                                  & \cellcolor[HTML]{DAE8FC}\textbf{85.14} & \cellcolor[HTML]{DAE8FC}\textbf{87.86} & \cellcolor[HTML]{DAE8FC}\textbf{85.60} \\ \hline
\rowcolor[HTML]{D0D0D0} 
\multicolumn{19}{|c|}{\cellcolor[HTML]{D0D0D0}Improved WideResNet classifier}                                                                                                                                                                                                                                                                                                                                                                                                                                                                                                                                                                                                                                                                                                                               \\ \hline
\cellcolor[HTML]{D0D0D0}RTSD + CGI-GAN     &                                        &                                        &                                        &                                        &                                        &                                       &                                        &                                        &                                        & 73.88                                  & 76.34                                  & 72.38                                  & 54.64                                  & 59.84                                  & 52.07                                  & 91.84                                  & 91.75                                  & 91.34                                  \\ \hline
\cellcolor[HTML]{D0D0D0}RTSD + Pasted      &                                        &                                        &                                        &                                        &                                        &                                       &                                        &                                        &                                        & 74.38                                  & 78.93                                  & 74.38                                  & 54.26                                  & 65.06                                  & 55.30                                  & 93.16                                  & 91.89                                  & 92.20                                  \\ \hline
\cellcolor[HTML]{D0D0D0}RTSD + Cycled      &                                        &                                        &                                        &                                        &                                        &                                       &                                        &                                        &                                        & 76.51                                  & 79.65                                  & 75.59                                  & \cellcolor[HTML]{DAE8FC}\textbf{58.75} & \cellcolor[HTML]{DAE8FC}\textbf{67.78} & \cellcolor[HTML]{DAE8FC}\textbf{58.73} & 93.10                                  & 90.74                                  & 91.33                                  \\ \hline
\cellcolor[HTML]{D0D0D0}RTSD + Styled      &                                        &                                        &                                        &                                        &                                        &                                       &                                        &                                        &                                        & \cellcolor[HTML]{DAE8FC}\textbf{76.69} & \cellcolor[HTML]{DAE8FC}\textbf{80.12} & \cellcolor[HTML]{DAE8FC}\textbf{76.24} & \cellcolor[HTML]{DAE8FC}\textbf{58.74} & 66.43                                  & 58.41                                  & \cellcolor[HTML]{DAE8FC}\textbf{93.45} & \cellcolor[HTML]{DAE8FC}\textbf{92.90} & \cellcolor[HTML]{DAE8FC}\textbf{92.89} \\ \hline
\end{tabular}
}\caption{Macro-averaged Precision, Recall and F1 for WideResNet classifiers trained on a mixture of real and synthetic data.} \label{tab:macrowrnetmetricmixture}
\end{table*}

\begin{table*}[h]
\resizebox{\textwidth}{!}{\begin{tabular}{|c|c|c|c|c|c|c|c|c|c|c|c|c|c|c|c|c|c|c|}
\hline
\rowcolor[HTML]{D0D0D0} 
\cellcolor[HTML]{D0D0D0}                   & \multicolumn{9}{c|}{\cellcolor[HTML]{D0D0D0}Metrics of softmax output}                                                                                                                                                                                                                                                                                                        & \multicolumn{9}{c|}{\cellcolor[HTML]{D0D0D0}Metrics with index from synthetic set}                                                                                                                                                                                                                                                                                             \\ \cline{2-19} 
\rowcolor[HTML]{D0D0D0} 
\cellcolor[HTML]{D0D0D0}                   & \multicolumn{3}{c|}{\cellcolor[HTML]{D0D0D0}all}                                                                         & \multicolumn{3}{c|}{\cellcolor[HTML]{D0D0D0}rare}                                                                        & \multicolumn{3}{c|}{\cellcolor[HTML]{D0D0D0}frequent}                                                                   & \multicolumn{3}{c|}{\cellcolor[HTML]{D0D0D0}all}                                                                         & \multicolumn{3}{c|}{\cellcolor[HTML]{D0D0D0}rare}                                                                        & \multicolumn{3}{c|}{\cellcolor[HTML]{D0D0D0}frequent}                                                                    \\ \cline{2-19} 
\rowcolor[HTML]{D0D0D0} 
\multirow{-3}{*}{\cellcolor[HTML]{D0D0D0}} & precision                              & recall                                 & F1                                     & precision                              & recall                                 & F1                                     & precision                             & recall                                 & F1                                     & precision                              & recall                                 & F1                                     & precision                              & recall                                 & F1                                     & precision                              & recall                                 & F1                                     \\ \hline
\rowcolor[HTML]{D0D0D0} 
\multicolumn{19}{|c|}{\cellcolor[HTML]{D0D0D0}Simple WideResNet classifier}                                                                                                                                                                                                                                                                                                                                                                                                                                                                                                                                                                                                                                                                                                                                 \\ \hline
\cellcolor[HTML]{D0D0D0}RTSD + CGI-GAN     & 40.45                                  & 46.19                                  & 38.15                                  & 22.30                                  & 39.96                                  & 24.32                                  & 57.40                                 & 52.01                                  & 51.07                                  & 36.56             & 37.52             & 31.72             & 17.71             & 31.30             & 18.27             & 54.17             & 43.34             & 44.27             \\ \hline
\cellcolor[HTML]{D0D0D0}RTSD + Pasted      & 49.11                                  & 66.45                                  & 50.82                                  & 28.35                                  & \cellcolor[HTML]{DAE8FC}\textbf{61.43} & 33.79                                  & 68.49                                 & 71.14                                  & 66.73                                  & 45.88             & 63.85             & 47.47             & 23.14             & 58.55             & 29.42             & 67.11             & 68.80             & 64.33             \\ \hline
\cellcolor[HTML]{D0D0D0}RTSD + Cycled      & 51.95                                  & 65.27                                  & 53.29                                  & \cellcolor[HTML]{DAE8FC}\textbf{31.40} & 56.33                                  & \cellcolor[HTML]{DAE8FC}\textbf{35.73} & 71.15                                 & 73.62                                  & 69.69                                  & 49.71             & 66.33             & 51.81             & 28.10             & \cellcolor[HTML]{DAE8FC}\textbf{59.23} & 33.39             & 69.90             & 72.95             & 69.01             \\ \hline
\cellcolor[HTML]{D0D0D0}RTSD + Styled      & \cellcolor[HTML]{DAE8FC}\textbf{53.31} & \cellcolor[HTML]{DAE8FC}\textbf{68.97} & \cellcolor[HTML]{DAE8FC}\textbf{54.96} & 30.51                                  & 59.86                                  & \cellcolor[HTML]{DAE8FC}\textbf{35.72} & \cellcolor[HTML]{DAE8FC}\textbf{74.60} & \cellcolor[HTML]{DAE8FC}\textbf{77.48} & \cellcolor[HTML]{DAE8FC}\textbf{72.93} & \cellcolor[HTML]{DAE8FC}\textbf{53.70}  & \cellcolor[HTML]{DAE8FC}\textbf{68.80} & \cellcolor[HTML]{DAE8FC}\textbf{56.14} & \cellcolor[HTML]{DAE8FC}\textbf{30.95} & 58.57             & \cellcolor[HTML]{DAE8FC}\textbf{36.62} & \cellcolor[HTML]{DAE8FC}\textbf{74.95} & \cellcolor[HTML]{DAE8FC}\textbf{78.36} & \cellcolor[HTML]{DAE8FC}\textbf{74.37} \\ \hline
\rowcolor[HTML]{D0D0D0} 
\multicolumn{19}{|c|}{\cellcolor[HTML]{D0D0D0}Improved WideResNet classifier}                                                                                                                                                                                                                                                                                                                                                                                                                                                                                                                                                                                                                                                                                                                               \\ \hline
\cellcolor[HTML]{D0D0D0}RTSD + CGI-GAN     &                                        &                                        &                                        &                                        &                                        &                                        &                                       &                                        &                                        & 45.73             & 55.96             & 45.42             & 26.32             & 48.38             & 29.94             & 63.85             & 63.04             & 59.87             \\ \hline
\cellcolor[HTML]{D0D0D0}RTSD + Pasted      &                                        &                                        &                                        &                                        &                                        &                                        &                                       &                                        &                                        & 55.60             & 73.37             & 58.46             & 35.80             & 66.02             & 41.67             & 74.08             & 80.24             & 74.15             \\ \hline
\cellcolor[HTML]{D0D0D0}RTSD + Cycled      &                                        &                                        &                                        &                                        &                                        &                                        &                                       &                                        &                                        & 56.80             & 69.80              & 58.65             & 36.75             & 60.69             & 41.42             & 75.53             & 78.31             & 74.74             \\ \hline
\cellcolor[HTML]{D0D0D0}RTSD + Styled      &                                        &                                        &                                        &                                        &                                        &                                        &                                       &                                        &                                        & \cellcolor[HTML]{DAE8FC}\textbf{58.79} & \cellcolor[HTML]{DAE8FC}\textbf{74.00} & \cellcolor[HTML]{DAE8FC}\textbf{61.03} & \cellcolor[HTML]{DAE8FC}\textbf{39.13} & \cellcolor[HTML]{DAE8FC}\textbf{66.38} & \cellcolor[HTML]{DAE8FC}\textbf{43.69} & \cellcolor[HTML]{DAE8FC}\textbf{77.15} & \cellcolor[HTML]{DAE8FC}\textbf{81.12} & \cellcolor[HTML]{DAE8FC}\textbf{77.23} \\ \hline
\end{tabular}
}\caption{Macro-averaged Precision, Recall and F1 measures for WideResNet classifiers trained only on synthetic samples.} \label{tab:macrowrnetmetriconly}
\end{table*}


The best results are highlighted in tables. Obtained values show that approaches
$Cycled$ and $Styled$ compete in terms of quality for target classifiers. It’s
hard to say which data is better. It depends on the specific task in which
target classifiers will be used. However, $94.11\%$ is the best accuracy value
that was obtained by classifying both rare and frequent classes with an improved
classifier (table \ref{tab:betterwideresnetmix}). It was achieved by training using the $Styled$ data and
classification by index from corresponding synthetic data.Previously, the best quality was $93.52\%$. Micro-average recall of rare traffic signs has also greatly improved from $70.16$ to $76.33$. Table \ref{tab:macrowrnetmetricmixture} also shows that we were able to improve macro-averaged precision, recall and F1 for the simple and improved WideResNet classifiers using the proposed synthetic data in comparison with the $CGI-GAN$. The best results were shown by methods $Cycled$ and $Styled$. For all classes, F1-measure has grown from $72.38$ to $76.24$ with $Styled$, for rare from $52.07$ to $58.73$ with $Cycled$, and for frequent ones from $91.34$ to $92.89$ with $Styled$.

Improved classifier still allows obtaining better quality compared to the usual
one (tables \ref{tab:betterwideresnetmix}, \ref{tab:macrowrnetmetricmixture}). For usual maximum accuracy, the value is $92.82\%$ (this is less than $94.11\%$
for the improved one). Macro-averaged precision, recall and F1 also better with improved classifier than with usual. This once again confirms the assumption of previous article \cite{betterwideresnet}.

It is also seen that proposed synthetic data significantly improve the quality of
classification when training only on synthetic data. Previously, the best accuracy
was $60.55\%$, and now $73.03\%$ (table \ref{tab:betterwideresnetonly}).

Therefore we conclude that usage of proposed synthetic samples during training
the process allows improving the quality of the WideResNet classifier.

\subsection{Detector results}
Next, we present the results of experiments with a detector. Table
\ref{tab:detectormix} shows AUC values for a detector trained on a mixture of
real and synthetic samples. The table \ref{tab:detectoronly} shows AUC
measurements for a detector trained only on synthetic samples.

\begin{table*}[h]
\resizebox{\textwidth}{!}{\begin{tabular}{|
>{\columncolor[HTML]{C0C0C0}}c |c|c|c|c|c|c|}
\hline
\multicolumn{1}{|l|}{\cellcolor[HTML]{C0C0C0}} & \multicolumn{3}{c|}{\cellcolor[HTML]{C0C0C0}Without classifier}                                                          & \multicolumn{3}{c|}{\cellcolor[HTML]{C0C0C0}With classifier}                                                             \\ \cline{2-7}
\multicolumn{1}{|l|}{\cellcolor[HTML]{C0C0C0}} & all                                    & rare                                   & frequent                               & all                                    & rare                                   & frequent                               \\ \hline
RTSD                                           & 89.09                                  & 85.86                                  & 89.25                                  & 86.01                                  & 58.56                                  & 86.61                                  \\ \hline
RTSD + Cgi                                     & 88.56                                  & 85.72                                  & 88.72                                  & 83.84                                  & 48.51                                  & 85.15                                  \\ \hline
RTSD + Inpaint                                 & 88.61                                  & 86.63                                  & 88.71                                  & 76.41                                  & 34.00                                  & 82.93                                  \\ \hline
RTSD + Pasted                                  & 88.98                                  & 86.59                                  & 89.09                                  & 85.81                                  & 59.98                                  & 86.40                                  \\ \hline
RTSD + Cycled                                  & 88.98                                  & 86.29                                  & 89.13                                  & 86.11                                  & 60.13                                  & 86.62                                  \\ \hline
RTSD + Styled                                  & 89.01                                  & \cellcolor[HTML]{DAE8FC}\textbf{86.78} & 89.13                                  & 85.39                                  & 64.20                                  & 86.13                                  \\ \hline
RTSD + KDE-only-synt                         & 89.03                                  & 86.34                                  & 89.17                                  & 85.59                                  & 63.83                                  & 86.28                                  \\ \hline
RTSD + NN-only-synt                          & 88.79                                  & 86.34                                  & 88.92                                  & 85.43                                  & 64.76                                  & 86.11                                  \\ \hline
RTSD + KDE-manystyled                         & 88.88                                     & 86.32                                     & 89.01                                     & 85.34                                     & 62.70                                     & 86.01                                     \\ \hline
RTSD + NN-manyStyled                          & 88.95                                  & 86.18                                  & 89.09                                  & 85.37                                  & 63.65                                  & 86.07                                  \\ \hline
RTSD + KDE-additional                         & 89.11                                  & 86.52                                  & 89.22                                  & 85.99                                  & 64.83                                  & 86.54                                  \\ \hline
RTSD + NN-additional                          & \cellcolor[HTML]{DAE8FC}\textbf{89.17} & 86.62                                  & \cellcolor[HTML]{DAE8FC}\textbf{89.31} & \cellcolor[HTML]{DAE8FC}\textbf{86.16} & \cellcolor[HTML]{DAE8FC}\textbf{64.96} & \cellcolor[HTML]{DAE8FC}\textbf{86.70} \\ \hline
\end{tabular}
}\caption{Detector trained on a mixture of a real and synthetic samples.} \label{tab:detectormix}
\end{table*}

\begin{table*}[h]
\resizebox{\textwidth}{!}{\begin{tabular}{|
>{\columncolor[HTML]{C0C0C0}}c |c|c|c|c|c|c|}
\hline
\multicolumn{1}{|l|}{\cellcolor[HTML]{C0C0C0}} & \multicolumn{3}{c|}{\cellcolor[HTML]{C0C0C0}Without classifier}                                                          & \multicolumn{3}{c|}{\cellcolor[HTML]{C0C0C0}With classifier}                                                             \\ \cline{2-7}
\multicolumn{1}{|l|}{\cellcolor[HTML]{C0C0C0}} & all                                    & rare                                   & frequent                               & all                                    & rare                                   & frequent                               \\ \hline
only CGI                                       & 10.70                                  & 13.23                                  & 10.63                                  & 8.81                                   & 8.53                                   & 8.81                                   \\ \hline
only Inpaint                                   & 55.23                                  & \cellcolor[HTML]{DAE8FC}\textbf{55.26} & 56.26                                  & 15.89                                  & 13.79                                  & 15.96                                  \\ \hline
only Pasted                                    & 38.22                                  & 38.97                                  & 38.17                                  & 19.85                                  & 18.80                                  & 19.91                                  \\ \hline
only Cycled                                    & 37.20                                  & 42.97                                  & 37.13                                  & 25.88                                  & 24.42                                  & 25.94                                  \\ \hline
only Styled                                    & \cellcolor[HTML]{DAE8FC}\textbf{62.12} & 54.26                                  & \cellcolor[HTML]{DAE8FC}\textbf{62.36} & \cellcolor[HTML]{DAE8FC}\textbf{39.99} & \cellcolor[HTML]{DAE8FC}\textbf{32.51} & \cellcolor[HTML]{DAE8FC}\textbf{40.35} \\ \hline
only KDE-only-synt                           & 51.22                                  & 42.76                                  & 51.61                                  & 32.36                                  & 32.68                                  & 25.83                                  \\ \hline
only NN-only-synt                            & 50.49                                  & 42.02                                  & 50.83                                  & 31.88                                  & 25.52                                  & 32.19                                  \\ \hline
only KDE-manystyled                           & 60.39                                     & 52.99                                     & 60.68                                     & 39.09                                     & 32.59                                     & 39.44                                     \\ \hline
only NN-manystyled                            & 60.33                                  & 52.57                                  & 60.64                                  & 38.66                                  & 31.63                                  & 39.04                                  \\ \hline
\end{tabular}
}\caption{Detector trained on synthetic samples.} \label{tab:detectoronly}
\end{table*}

The best results are highlighted in tables. It can be seen that without
intelligent placement with the neural network it is not possible to improve
the detection quality of frequent classes by any synthetic data. It is $89.25$ for
frequent classes when training only on real data. The best quality using
synthetic samples without neural network placement is achieved at $Styled$ and
is equal to $89.13$. On rare signs, AUC increases from $85.86$ to $86.78$.
It can be seen from the table that proposed approaches for post-processing of synthetic signs work better than already existed. At the same time, we improved the quality of sequential detection and classification of traffic signs due to classifiers' improvement. Our average AUC increased from $86.01$ to $86.11$. On rare signs, the increase is more
significant -- from $58.56$ to $64.20$.

The best results are obtained if VAE-based neural network is used for placement of new signs. Gaussian kernel density estimation works a bit worse. That means
that the proposed method for location generation with a neural network is better for
synthetic training data than random placement with samples generated from
distribution. Without a classifier, we achieved $89.17$ average AUC and $89.31$ for
frequent signs which is the best result. For rare classes metric is a bit worse
($86.62$) than achieved with $Styled$ $86.78$. When using $Styled$ classifier
this method has the best results for average and for rare/frequent. Thus our best
average AUC is $86.16$.

Training only on synthetic data shows that the synthetic set closest to real data is
$Styled$. It turns out to achieve a significantly higher quality of detection in
comparison with other synthetic sets. On average it equals to $62.12$ without a
classifier and $39.99$ with a classifier.

\section{Conclusion}
In this paper, we proposed a neural network-based method for embedding new
synthetic traffic signs in road images. The proposed method consists of networks for
the placement and processing of new signs. For placement, we use a neural network with
Spatial Transformer Network to predict the best locations for additional signs. Also,
we developed three improved models for the processing of traffic signs. The idea of
the proposed approaches is to use neural networks consisting of two parts. One is
used to inpaint traffic signs that have been already located in the image, and the
second is used to embed new ones. It the two of the proposed methods these two parts are being trained jointly, but in the third approach separately.

We also made a comparison of different synthetic data quality by training target
neural networks. It showed that our method for generating synthetic training
samples improved quality for detector and classifier of traffic signs. For all
sign classes, recognition quality improved. The most noticeable improvement was
achieved for rare classes which absolutely absent in the original real training set.

{\small
\bibliographystyle{ieeetr}
\bibliography{egbib}
}

\end{document}